\documentclass[11pt]{article}
\usepackage{enumitem}

\usepackage{acl}
\usepackage{times}
\usepackage{latexsym}
\usepackage{amsmath}
\usepackage{amssymb}
\usepackage[T1]{fontenc}
\usepackage[utf8]{inputenc}
\usepackage{microtype}
\usepackage{graphicx}
\usepackage{booktabs}
\usepackage[table]{xcolor}
\usepackage{listings}
\usepackage[most]{tcolorbox}
\usepackage{algorithm}
\usepackage{algpseudocode}
\usepackage{float}
\usepackage{placeins}

\graphicspath{{figures/}}

\definecolor{promptCardBg}{HTML}{F7F7F9}
\definecolor{promptCardFrame}{HTML}{B8BDC7}
\definecolor{promptCardHeader}{HTML}{3F4754}
\definecolor{definitionBoxBg}{HTML}{F8FAFC}
\definecolor{definitionBoxFrame}{HTML}{64748B}

\lstdefinestyle{appendixprompt}{
    basicstyle=\scriptsize\ttfamily,
    breaklines=true,
    breakatwhitespace=false,
    columns=fullflexible,
    keepspaces=true,
    frame=none,
    showstringspaces=false
}

\newtcblisting{promptcard}[1][]{
    enhanced,
    listing only,
    listing engine=listings,
    colback=promptCardBg,
    colframe=promptCardFrame,
    colbacktitle=promptCardHeader,
    coltitle=white,
    fonttitle=\bfseries,
    boxrule=0.35pt,
    arc=2pt,
    left=5pt,
    right=5pt,
    top=5pt,
    bottom=5pt,
    before skip=0.5em,
    after skip=0.75em,
    listing options={style=appendixprompt},
    #1
}

\newtcolorbox{definitionbox}[1][]{
    enhanced,
    breakable,
    colback=definitionBoxBg,
    colframe=definitionBoxFrame,
    boxrule=0.35pt,
    arc=2pt,
    left=5pt,
    right=5pt,
    top=5pt,
    bottom=5pt,
    before skip=0.5em,
    after skip=0.75em,
    #1
}

\title{Low-Resource Safety Failures Are Action Failures,\\ Not Representation Failures}

\author{
Rashad Aziz \quad
Ikhlasul Akmal Hanif \quad
Fajri Koto \\
Mohamed bin Zayed University of Artificial Intelligence \\
\texttt{\small \{rashad.aziz,ikhlasul.hanif,fajri.koto\}@mbzuai.ac.ae}
}

\begin{document}
\maketitle


\begin{abstract}
Language models often answer harmful requests in low-resource languages (LRLs) that they refuse in high-resource languages (HRLs). Across three instruction-tuned models and 23 languages, harmful refusal falls from 87.9\% in HRLs to 43.9\% in LRLs, while harmless refusal remains low. A common explanation is that models represent harmfulness weakly in LRLs. We test whether harmfulness is instead represented but does not reliably produce refusal. Across three models, a harmfulness direction learned from HRL activations still separates harmful from harmless LRL prompts, showing that complete absence of harmfulness information cannot explain many failures. However, harmfulness scores shift downward for LRL prompts, making harmful prompts less likely to reach the range associated with refusal. Motivated by this shift, we train a low-rank logistic classifier on HRL activations and calibrate its threshold with a few target-language examples. During generation, the classifier conditionally adds or ablates the HRL harmfulness direction. With the same HRL data and 32 target-language examples per class, CAST remains limited by low harmful refusal and AdaSteer by high harmless refusal, yielding mean refusal selectivity ($\Delta$ = harmful $-$ harmless refusal) of 33.6 and 6.8, respectively. Our intervention reaches 54.5 while preserving MMLU utility. HRL-only calibration improves selectivity for Qwen and Gemma, whereas Llama benefits from target-language calibration. These results show that recalibrating existing representations can offer a training-free method for repairing low-resource safety failures.

\end{abstract}

\begin{figure}[t]
    \centering
    \includegraphics[width=0.8\columnwidth]{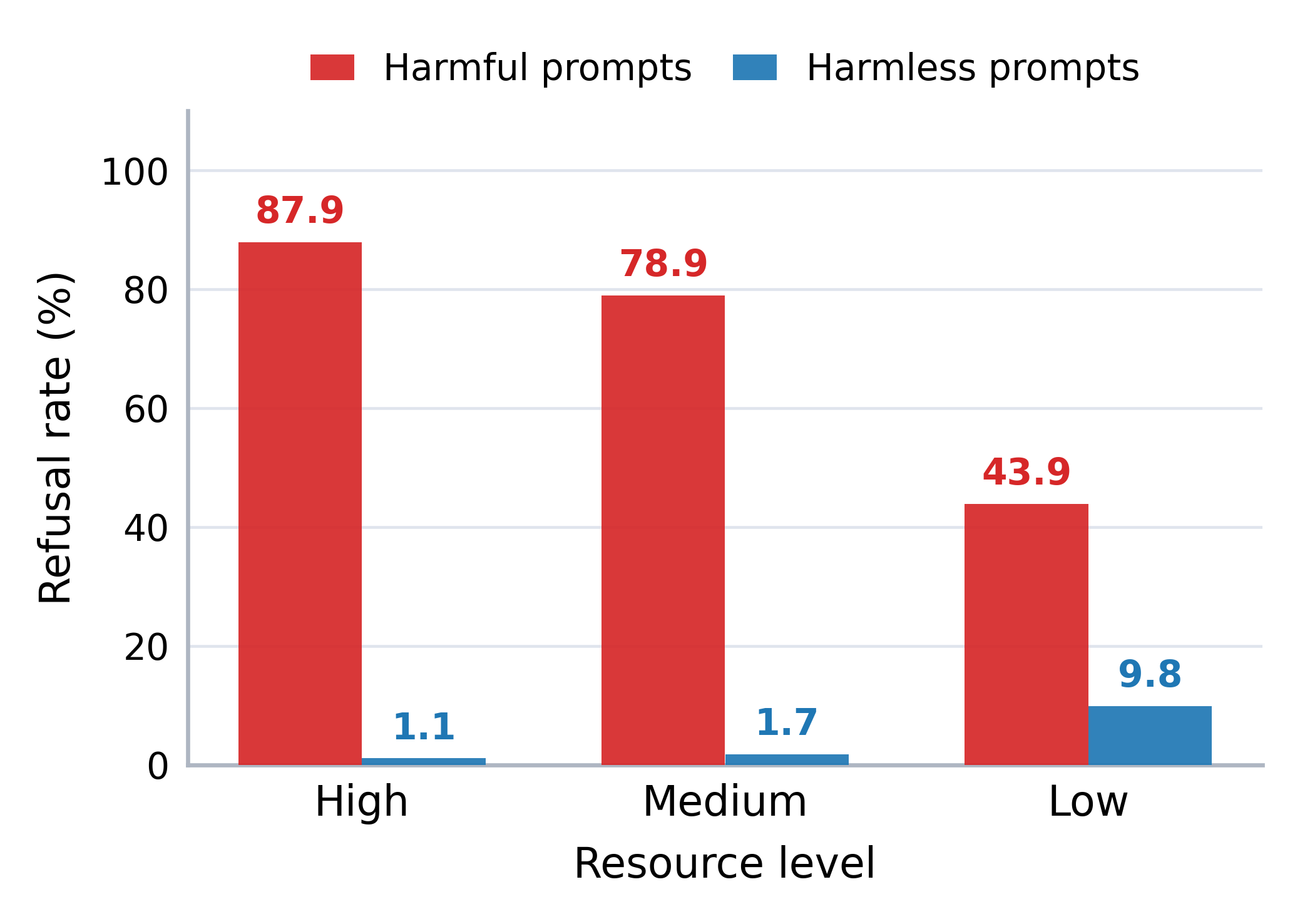}
    \caption{
        \textbf{Refusal degrades with lower language resource tier.}
        Macro refusal rates across Qwen2.5-7B-Instruct, Gemma-2-9B-it, and Llama-3.1-8B-Instruct show harmful refusal falling sharply in LRLs while harmless refusal stays low.
    }
    \label{fig:motivating-refusal-gap}
\end{figure}

\section{Introduction}
 

LLMs are trained to refuse harmful instructions while answering harmless ones \cite{ouyang2022instructgpt,bai2022constitutional}. Yet this behavior is uneven across languages: harmful prompts in lower-resource languages often succeed where equivalent English prompts are refused \cite{deng2023multijail,yong2023lowresource,wang-etal-2024-languages,shen-etal-2024-language}. Figure~\ref{fig:motivating-refusal-gap} illustrates this pattern, showing that refusal rates for harmful prompts drop sharply as language resources decrease, while harmless prompts remain largely unaffected. These failures are typically measured behaviorally, leaving their internal cause unclear.


When a model answers a harmful low-resource request, two failures are observationally indistinguishable. The model may fail to recognize the request as harmful, or it may \textit{recognize} it but fail to \textit{act} on that recognition. Prior work has documented the behavioral gap and proposed mitigations that transfer a stronger safety signal from English or other high-resource languages \cite{zhang-etal-2025-english,zhao-etal-2025-mpo,bu2026alignonce,yang-etal-2026-lasa}, but the underlying question has remained open: \textit{do lower-resource prompts lack a usable harmfulness representation} \cite{wang2025refusaluniversal,verma2025hiddenspace}, \textit{or do they fail to use one that is already present?}


We first test whether harmfulness remains encoded in the residual stream, where directions can affect refusal \cite{arditi2024refusal,wang2025refusaluniversal}. If a direction learned from HRL activations does not separate harmful from harmless low-resource prompts, weaker harmfulness representations could explain the gap. If it remains discriminative when refusal fails, harmfulness is encoded but does not consistently produce refusal.


Across a 23-language PolyRefuse extension \cite{wang2025refusaluniversal}, harmfulness remains linearly separable in low-resource activations. Ablating the HRL harmfulness direction from the residual stream reduces refusal, while adding it recovers refusal in many low-resource settings. Lower-resource prompts often have smaller projections, so more of the direction must be added to produce refusal. However, some prompts still remain unrefused even after we add the direction.


We therefore introduce a low-rank logistic classifier, which we call a few-shot latent gate, and calibrate it with a small target-language set. The method adds the HRL harmfulness direction to prompts classified as harmful and ablates it from prompts classified as harmless. It improves selective refusal across most model--language pairs and preserves benign instruction following better than adapted steering baselines \cite{lee2025cast,zhao-etal-2025-adasteer}.

Our contributions can be summarized as follows:
\begin{itemize}[noitemsep, topsep=0pt]
    \item We show that complete absence of harmfulness information cannot explain many low-resource refusal failures: harmfulness remains decodable from the HRL direction when refusal fails.
    \item We distinguish weak harmful--harmless separation, errors by our learned classifier, and non-refusals after direction addition, and show that their prevalence differs across models.
    \item We propose a few-shot latent gate that improves selective refusal and preserves benign instruction following better than adapted baselines under matched supervision; \href{https://github.com/rashadaziz/low-resource-safety}{code is available on github}.
\end{itemize}




\section{Related Work}
\paragraph{Multilingual Safety Gap.}
Prior work consistently shows that safety degrades in lower-resource languages \cite{deng2023multijail,yong2023lowresource,wang-etal-2024-languages,shen-etal-2024-language,yong-etal-2025-state}: unsafe generations increase as language resources decrease, translated AdvBench prompts elicit harmful outputs \cite{yong2023lowresource}, and non-English inputs yield more unsafe and irrelevant responses despite multilingual SFT and RLHF \cite{wang-etal-2024-languages,shen-etal-2024-language}. However, these findings are based on response-level behavior and do not explain the underlying mechanism. It remains unclear whether models fail to separate harmful from harmless prompts, or separate them but fail to translate this signal into refusal. In this work, we study this distinction at the representation level.

\paragraph{Multilingual Safety Alignment.}
Existing methods for closing the multilingual safety gap typically strengthen safety behavior through additional training or auxiliary safety mechanisms. Approaches include reward-gap optimization \cite{zhao-etal-2025-mpo}, representation consistency objectives \cite{bu2026alignonce,yang-etal-2026-lasa}, sparse or language-specific weight edits \cite{liang2026sparseweight,banerjee-etal-2025-soteria}, English-anchored multilingual alignment training \cite{zhang-etal-2025-english}, and multilingual safety classifiers \cite{kumar2025polyguard,verma-etal-2025-multiguard}. In this work, we take a complementary mechanistic approach. Rather than first adding new multilingual safety knowledge through weight updates or a separate text-level guard, we diagnose the gap in the residual stream and show that harmful--harmless information is already present in lower-resource activations. Our intervention then reuses this high-resource harmfulness signal and lightly recalibrates how it is routed into refusal, improving low-resource safety without updating model weights or training a new multilingual safety model.

\paragraph{Mechanistic Interpretability.}
LLMs often encode high-level concepts in latent directions \cite{zou2023representation,ying2026truthfulness,lu2026assistantaxis}. For refusal, \citet{arditi2024refusal} find a residual-stream direction whose removal suppresses refusal on harmful prompts and whose addition induces refusal on harmless prompts. Later work finds multi-dimensional refusal geometry \cite{wollschlager2025geometry,pan2025hiddendimensions,joad2026refusalmulti}, separable harmfulness and refusal concepts \cite{lee2025cast}, and token-position effects \cite{zhao2025harmfulness}. Multilingual interpretability also finds shared cross-lingual semantics \cite{wendler-etal-2024-llamas,dumas-etal-2025-separating,wu2025semantichub}, while \citet{wang2025refusaluniversal} show refusal directions transfer across safety-aligned languages. \citet{wang2025refusaluniversal,verma2025hiddenspace} posit that lower-resource safety gaps come from harmful--harmless overlap caused by weaker semantic understanding. We find this incomplete: lower-resource languages can encode nontrivial harmful--harmless separation that still fails to induce refusal.

\section{Setup: Models, Data, and Harmfulness Directions}

\paragraph{Languages and resource tiers.}
Because model-specific pretraining mixtures are usually unavailable, we use Common Crawl share as a proxy for resource level \citep{touvron2023llama,penedo2023refinedweb,gao2020pile,soldaini-etal-2024-dolma,wenzek-etal-2020-ccnet}: above 1\% for high-resource languages (HRLs), 0.1--1\% for medium-resource languages (MRLs), and below 0.1\% for low-resource languages (LRLs). The 10 HRLs are English, German, French, Spanish, Italian, Dutch, Polish, Russian, Chinese, and Japanese; the 7 MRLs are Arabic, Korean, Thai, Greek, Hebrew, Hindi, and Persian; and the 6 LRLs are Swahili, Amharic, Burmese, Khmer, Sinhala, and Yoruba.

\paragraph{Dataset.}
We extend\footnote{See Appendix~\ref{app:polyrefuse-details} for details.}
PolyRefuse \cite{wang2025refusaluniversal} to 23 languages: 10 HRLs,
7 MRLs, and 6 LRLs. Each language has 260/39/572 harmful train/validation/test prompts
and 200/200/500 harmless train/validation/test prompts. The train split is used
for our calibration and adapted AdaSteer; CAST additionally uses validation for
condition-layer and threshold tuning. The test split is solely used for
evaluation.

\paragraph{Models and refusal scoring.}
The main experiments use Qwen2.5-7B-Instruct, gemma-2-9b-it, and
Llama-3.1-8B-Instruct. We greedily decode harmful and harmless PolyRefuse
prompts and use LLM-as-a-judge \cite{liu-etal-2023-g,zheng2023judging} to label refusal.
We chose GPT-4o-mini as the judge for its broad multilingual coverage and cost efficiency.
In a 600-example human audit, the judge reaches 92.5\% agreement, refusal
F1 of 0.835, and Cohen's $\kappa$ of 0.787
(Appendix~\ref{app:human-validation-refusal-judge}). It produces more
false-positive refusals than false negatives, so absolute refusal and selectivity
values should be interpreted with this bias in mind. The representation AUC
analysis uses benchmark harmful/harmless labels and does not depend on the refusal judge.

\paragraph{Harmfulness directions.}
We use the standard contrastive-mean construction \cite{arditi2024refusal,wang2025refusaluniversal} to extract a one-dimensional encoding of harmfulness from model activations. For a prompt $x$, let $h_k(x) \in \mathbb{R}^d$ denote the residual-stream activation \citep{elhage2021mathematical} at the final token of layer $k$. For each language $\ell$, we average $h_k(x)$ separately over a balanced contrastive set of harmful prompts $\mathcal{D}_{\ell, H}$ and harmless prompts $\mathcal{D}_{\ell, N}$,
\[
\mu_{\ell, s} = \frac{1}{|\mathcal{D}_{\ell, s}|} \sum_{x \in \mathcal{D}_{\ell, s}} h_k(x), \quad s \in \{H, N\},
\]
and define the language-$\ell$ harmfulness direction as the $\ell_2$-normalized difference of means:
\[
v_\ell = \frac{\mu_{\ell, H} - \mu_{\ell, N}}{\|\mu_{\ell, H} - \mu_{\ell, N}\|_2}.
\]
By construction, $v_\ell$ points from the harmless cluster toward the harmful cluster, so projections $s(x) = h_k(x)^\top v_\ell$ are higher for harmful prompts than harmless ones. We call $s(x)$ the harmfulness score. We use the direction in two ways: its score measures whether harmfulness is decodable, while adding or removing the direction tests whether it changes refusal.


\begin{figure*}[t]
    \centering
    \includegraphics[width=0.8\textwidth]{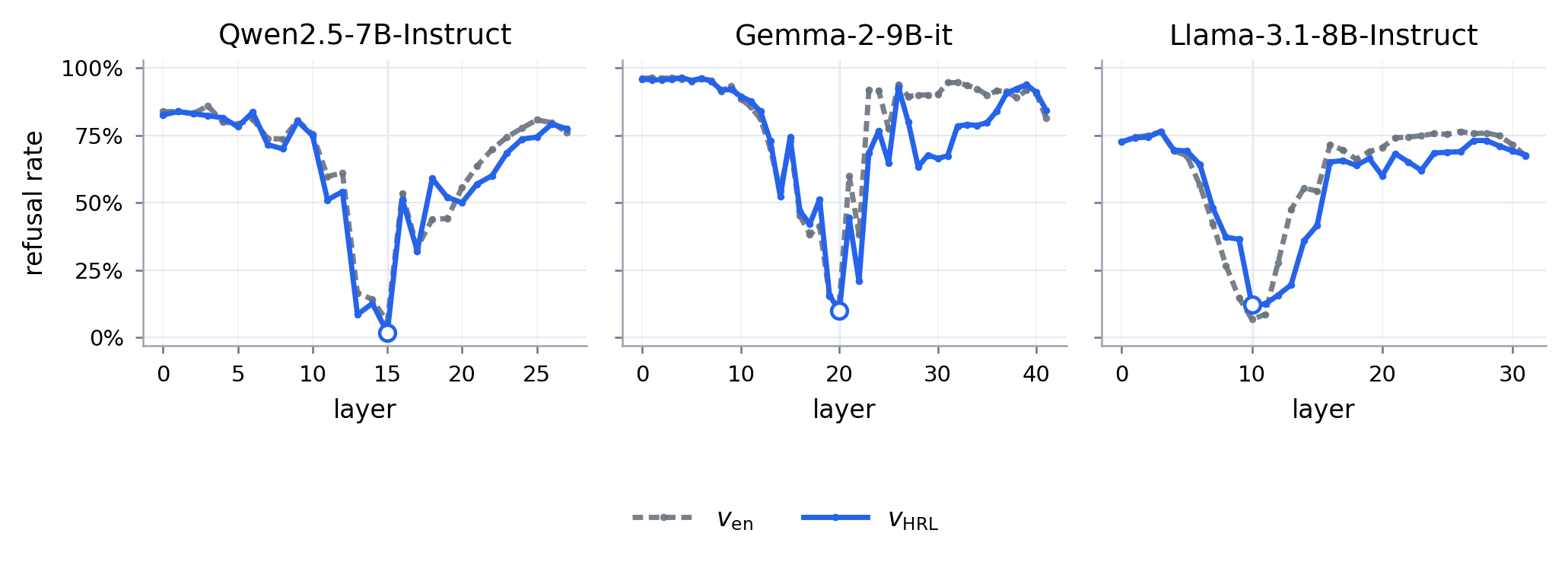}
    \caption{
        \textbf{Removing harmfulness directions reduces refusal in middle layers.}
        Removing the pooled HRL harmfulness direction reduces harmful refusal on HRL PolyRefuse prompts most at layer 15 for Qwen, 20 for Gemma, and 10 for Llama.
    }
    \label{fig:vhrl-ablation-sweep}
\end{figure*}

\section{Diagnosing the Multilingual Refusal Gap}
\label{sec:refusal-gap}


This section asks where multilingual safety alignment breaks down. We first measure the behavioral gap across resource tiers, then test whether the harmfulness representation responsible for refusal in high-resource languages is still present in lower-resource ones, and how that representation relates to the model's refusal behavior.

\paragraph{Harmful refusal degrades with resource tier.}
We generate greedy completions for harmful and harmless PolyRefuse prompts, score them with the refusal judge, and report refusal rates by language. Because a low-resource gap could reflect broad instruction-following degradation, we measure harmful and harmless refusal separately. Figure~\ref{fig:motivating-refusal-gap} summarizes the result, with Appendix Table~\ref{tab:main-tier-results} giving the per-language breakdown. Averaged across model families, harmful refusal falls from 87.9\% in HRLs to 78.9\% in MRLs and 43.9\% in LRLs. The pattern holds by model: Qwen drops from 85.6\% to 28.8\%, Llama from 83.5\% to 32.5\%, and the stronger Gemma-2-9B from 94.5\% to 70.3\%. Harmless refusal stays much lower throughout, with an aggregate LRL average of 9.8\%. The failure is therefore selective: models often answer harmful low-resource requests while still answering harmless ones. We call this the \emph{Multilingual Refusal Gap} and ask whether it reflects a missing harmfulness representation or a failure to act on one that is present.




\paragraph{Removing the HRL harmfulness direction reduces refusal.}
We test whether the HRL harmfulness direction affects refusal. Following \citet{arditi2024refusal}, we remove a candidate direction $v$ from the residual stream across all token positions, including during autoregressive generation:
\[
h_k(x) \leftarrow h_k(x) - v_\ell {v_\ell}^{T} h_k(x)
\]
If $v_\ell$ affects refusal, removing it at layer $k$ should reduce harmful
refusal. We therefore remove $v_{\mathrm{HRL}}$, the direction from pooled
activations of the 10 HRLs, at each layer and compare it to the English-derived
baseline $v_{en}$ \cite{arditi2024refusal}.

Figure~\ref{fig:vhrl-ablation-sweep} shows that removing $v_{\mathrm{HRL}}$ reduces harmful refusal to nearly zero in Qwen2.5-7B and below 15\% in Gemma-2-9B and Llama-3.1-8B at middle layers, while refusal stays much higher outside this band. Removing the pooled HRL direction reduces refusal more than removing $v_{\mathrm{en}}$ in Qwen and Gemma and by a similar amount in Llama. We define each model's \emph{extraction layer} as the layer where removing $v_{\mathrm{HRL}}$ most reduces harmful refusal: layer 15 for Qwen, 20 for Gemma, and 10 for Llama. Subsequent analyses use $v_{\mathrm{HRL}}$ at these layers.

\begin{figure}[!t]
    \centering
    \includegraphics[width=0.8\columnwidth]{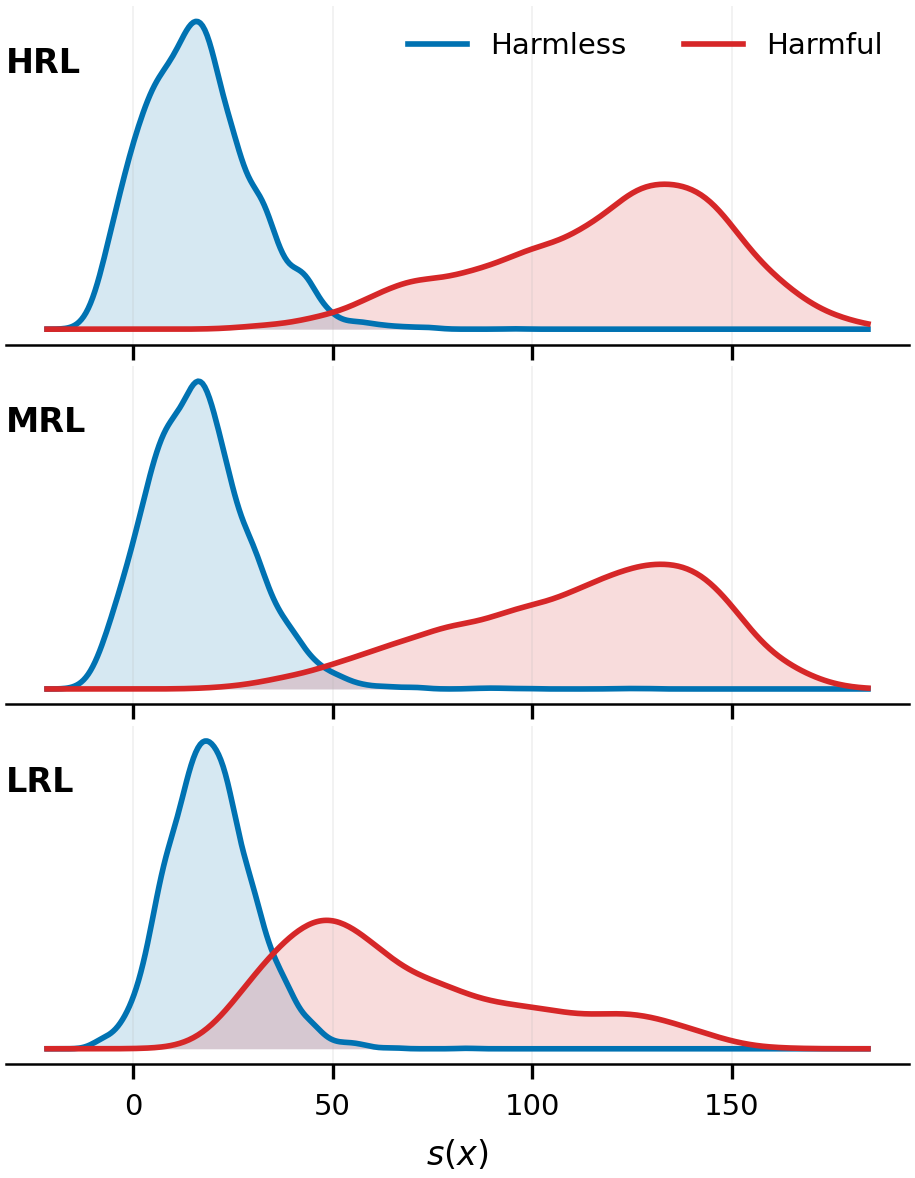}
    \caption{\textbf{Lower-resource prompts shift harmfulness-score distributions.} Gemma-2-9B projections on $v_{\mathrm{HRL}}$ show harmful and harmless $s(x)$ distributions within each resource tier; harmful projections shift downward most clearly in the LRL panel.}
    \label{fig:tier-projections}
\end{figure}

At the extraction layer, harmfulness scores $s(x) = h_k(x)^{\top} v_{\mathrm{HRL}}$ also separate harmful from harmless prompts (Figure~\ref{fig:tier-projections}; Appendix~\ref{fig:tier-projections-appendix}). Thus the same direction predicts harmfulness, and removing it reduces refusal.

\begin{figure}[!t]
    \centering
    \includegraphics[width=0.8\columnwidth]{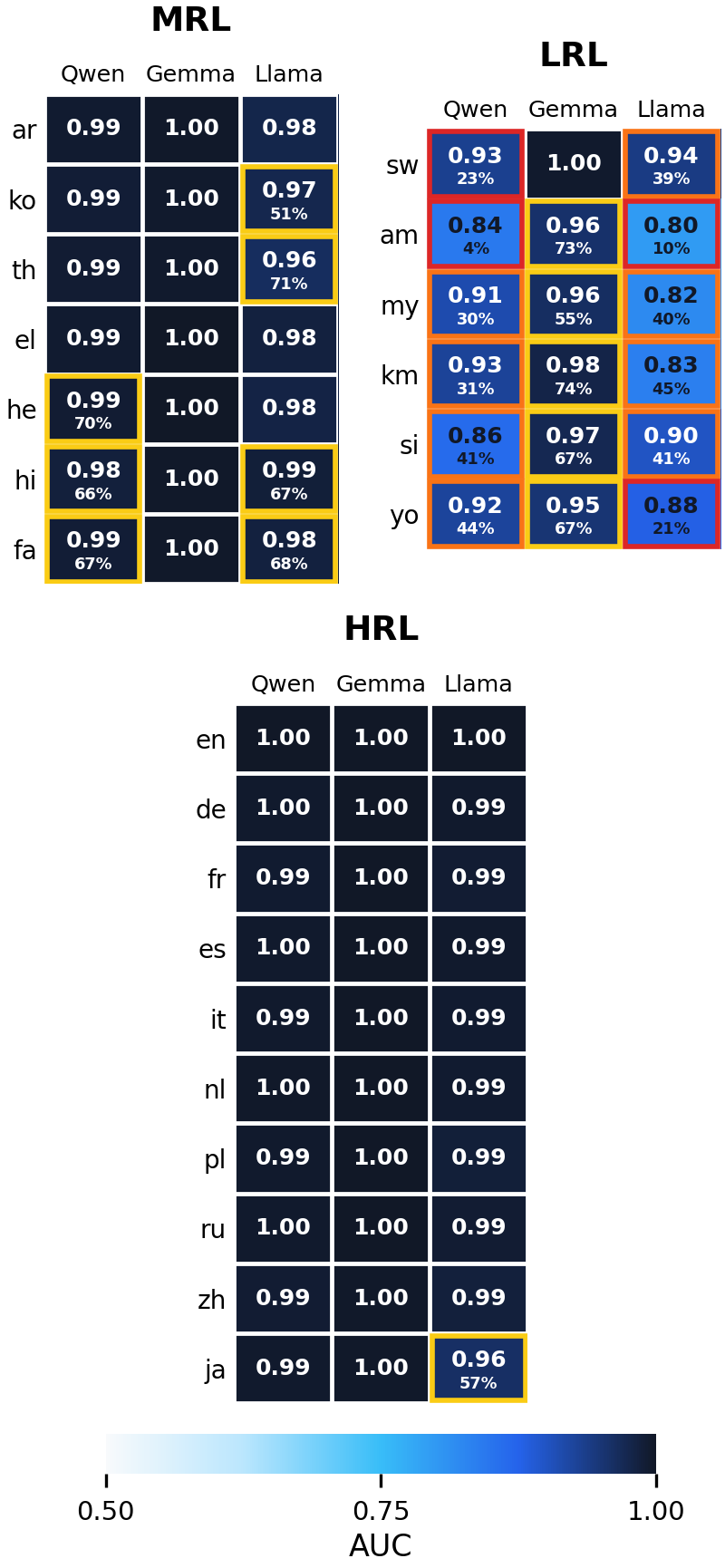}
    \caption{\textbf{Harmfulness remains decodable even when models fail to refuse.} Cells report PolyRefuse test AUC from $s(x)$ under the pooled HRL direction. Yellow, orange, and red borders mark harmful-refusal rates of 50--75\%, 25--50\%, and below 25\%, respectively. In highlighted cells, smaller numbers are actual harmful-refusal rates.}
    \label{fig:vhrl-auc}
\end{figure}

\paragraph{Harmfulness remains decodable in lower-resource languages.}
Prior work attributes the lower-resource refusal gap to weaker safety representations or understanding \cite{wang2025refusaluniversal,verma2025hiddenspace}. We test this directly by projecting LRL and MRL activations onto $v_{\mathrm{HRL}}$ and measuring how well the resulting harmfulness scores separate harmful from harmless prompts. Figure~\ref{fig:vhrl-auc} reports test AUC across all 23 languages: even the weakest LRL cells remain far above chance, with most languages above 0.85. Several cells combine high AUC with low refusal. Harmfulness is therefore decodable from the same direction whose removal reduces HRL refusal, even when the model does not refuse.

\begin{figure}[!t]
    \centering
    \includegraphics[width=0.8\columnwidth]{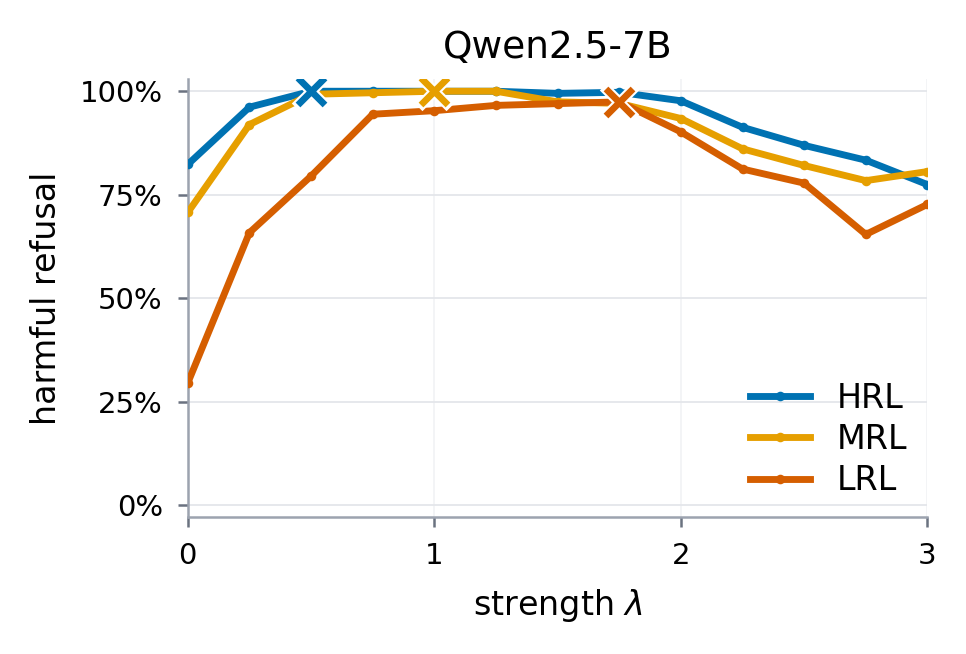}
    \caption{\textbf{Adding the HRL direction can recover refusal, but the needed amount differs by tier.} Curves average Qwen2.5-7B harmful refusal within each resource tier after adding $\lambda v_{\mathrm{HRL}}$. Crosses mark tier-wise peaks.}
    \label{fig:refusal-activation-sweep}
\end{figure}

\paragraph{Low-resource projections are smaller in magnitude.}
Harmfulness scores change across resource tiers even when harmful and harmless prompts remain separable. Figure~\ref{fig:tier-projections} shows Gemma's $s(x)$ distributions within each resource tier: in the aggregated LRL panel, harmful and harmless prompts remain separable, but the mean harmful score shifts downward. Qwen and Llama show the same pattern (Figure~\ref{fig:tier-projections-appendix}). We test whether this shift helps explain the refusal failure by adding $v_{\mathrm{HRL}}$ with increasing strength:
\[
\begin{aligned}
h_k(x) &\leftarrow h_k(x) + \lambda v_{\mathrm{HRL}}, \\
\lambda &\in \{0, 0.25, 0.5, \ldots, 3\}.
\end{aligned}
\]
Figure~\ref{fig:refusal-activation-sweep} shows that adding $v_{\mathrm{HRL}}$ recovers LRL refusal, but lower-resource tiers need larger $\lambda$ to reach peak refusal than HRLs do. Thus the amount added must account for the shift in harmfulness scores. Some settings may still require new representation learning when harmful--harmless separation is weak.

\begin{figure*}[!t]
   \centering
   \includegraphics[width=0.8\textwidth]{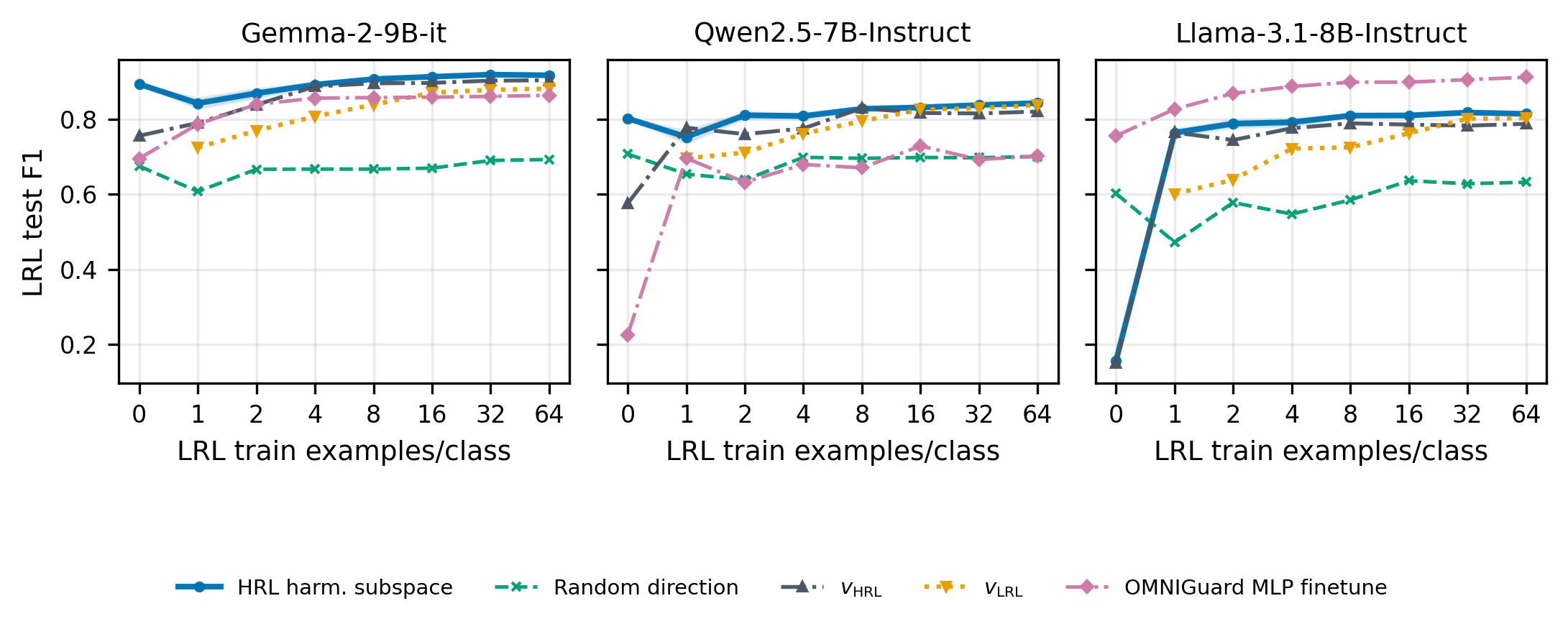}
   \caption{
   \textbf{A few target-language examples calibrate harmfulness classification in LRLs.}
   Each curve reports macro LRL test F1 after choosing a language-specific
   binary threshold with \(b\) harmful and \(b\) harmless target-language
   examples. The random-direction curve is a one-dimensional control whose
   threshold is calibrated with the same target-language examples.
   }
   \label{fig:fewshot-gates}
\end{figure*}

\section{A Few-Shot Latent Gate for Conditional Steering}
\subsection{Calibrating a Few-Shot Latent Gate}
If low-resource harmfulness is present but shifted, the next question is
whether it can be used with only minimal target-language supervision. We
therefore test whether an HRL-aligned harmfulness readout becomes a reliable 
LRL classifier after seeing just a few harmful and harmless examples.

\paragraph{Few-shot threshold calibration.}
Our classifier maps a harmfulness score to a binary label using a learned threshold. Direction scores use \(s(x)=h_k(x)^\top v\); subspace scores first
project \(h_k(x)\), then train logistic regression on the projected coordinates
(Algorithm~\ref{alg:hrl-subspace-gate}), using the pre-sigmoid logit as
\(s(x)\). For each budget and LRL, we choose the threshold \(\tau_\ell\) that
maximizes F1 on \(b\) harmful and \(b\) harmless training examples. When
\(b=0\), we calibrate on HRL training data. Thresholds are frozen before
held-out LRL evaluation, and test F1 is averaged across LRLs. All experiments in
this section were repeated on 10 seeds.

\paragraph{Few target-language examples calibrate the HRL direction.}
Figure~\ref{fig:fewshot-gates} compares fixed \(v_{\mathrm{HRL}}\) against a
target-language direction \(v_{\mathrm{LRL}}\) estimated from the same examples.
Zero-shot HRL thresholds can be poorly calibrated, especially for Qwen and Llama,
but one or two target-language examples per class often improve held-out F1.
Thus \(v_{\mathrm{HRL}}\) is more sample-efficient than estimating
\(v_{\mathrm{LRL}}\) from scratch: local directions improve with more labels, but
at small budgets are usually worse than LRL-specific calibration of the HRL
direction. The HRL direction still separates harmful from harmless LRL prompts,
but its threshold shifts.

\begin{algorithm}[t]
\small
\caption{Training the HRL harmfulness subspace classifier}
\label{alg:hrl-subspace-gate}
\begin{algorithmic}[1]
\Require Layer-$k$ activations $h_k(\cdot)$, rank $r$, target-language budget $b$

\Statex \textbf{Step 1: Compute per-language harmfulness directions.}
\Statex \hspace{1em} $\mu_{\ell,H}$, $\mu_{\ell,N}$: mean activations of harmful (H) and harmless (N) prompts in language $\ell$.
\For{each source HRL $\ell$}
    \State $d_\ell \gets \mathrm{normalize}\!\left(\mu_{\ell,H}^{\mathrm{train}} - \mu_{\ell,N}^{\mathrm{train}}\right)$
\EndFor

\Statex \textbf{Step 2: Build rank-$r$ subspace from pooled HRL directions.}
\State \(B_r \gets\) top \(r\) right-singular vectors of row stack
\([d_\ell]_{\ell\in\mathrm{HRL}}\)
\State $z(x) \gets h_k(x)^\top B_r$ \Comment{subspace features}

\Statex \textbf{Step 3: Assemble training set $\mathcal{T}$.}
\If{$b > 0$}
    \State Sample $b$ harmful and $b$ harmless target-language examples
    \State $\mathcal{T} \gets$ HRL train examples $\cup$ sampled target examples
\Else
    \State $\mathcal{T} \gets$ HRL train examples
\EndIf
\State Standardize $z(x)$ using statistics from $\mathcal{T}$

\Statex \textbf{Step 4: Fit classifier and select threshold.}
\State Fit logistic regression $g(z) = w^\top z + c$ on $\mathcal{T}$ (L2-regularized, balanced cross-entropy)
\State Select threshold $\tau$ on training logits ($\mathcal{T}$)
\State Freeze \(B_r\), \(g\), and the threshold; evaluate classifier F1 on the target test split
\end{algorithmic}
\end{algorithm}

\paragraph{The low-rank classifier is competitive with higher-capacity guards.}
The same pattern (Figure~\ref{fig:fewshot-gates}) holds against higher-capacity OMNIGuard fine-tuning
\cite{verma-etal-2025-multiguard}. To match source data, OMNIGuard is trained on HRLs
and fine-tuned separately for each LRL with the same \(b\) examples per class. It
improves with target-language examples, but even against the higher-capacity OMNIGuard baseline,
the low-rank classifiers remain competitive or stronger, especially for Qwen.

\paragraph{A shared HRL subspace improves the classifier.}
We also replace the single HRL direction with a rank-10 harmfulness subspace fit
from HRL training languages (Algorithm~\ref{alg:hrl-subspace-gate}). The subspace
is still HRL-only before calibration, but it captures multiple HRL harmfulness
axes, allowing the classifier to use more than one high-resource direction. Against a random-direction
control calibrated with the same examples, the learned HRL subspace (Figure~\ref{fig:fewshot-gates})
wins at every nonzero budget for all three models, with the largest gap for Gemma. The gain
therefore reflects harmfulness-aligned geometry beyond threshold calibration.

\begin{figure}[!t]
    \centering
    \includegraphics[width=\columnwidth,keepaspectratio]{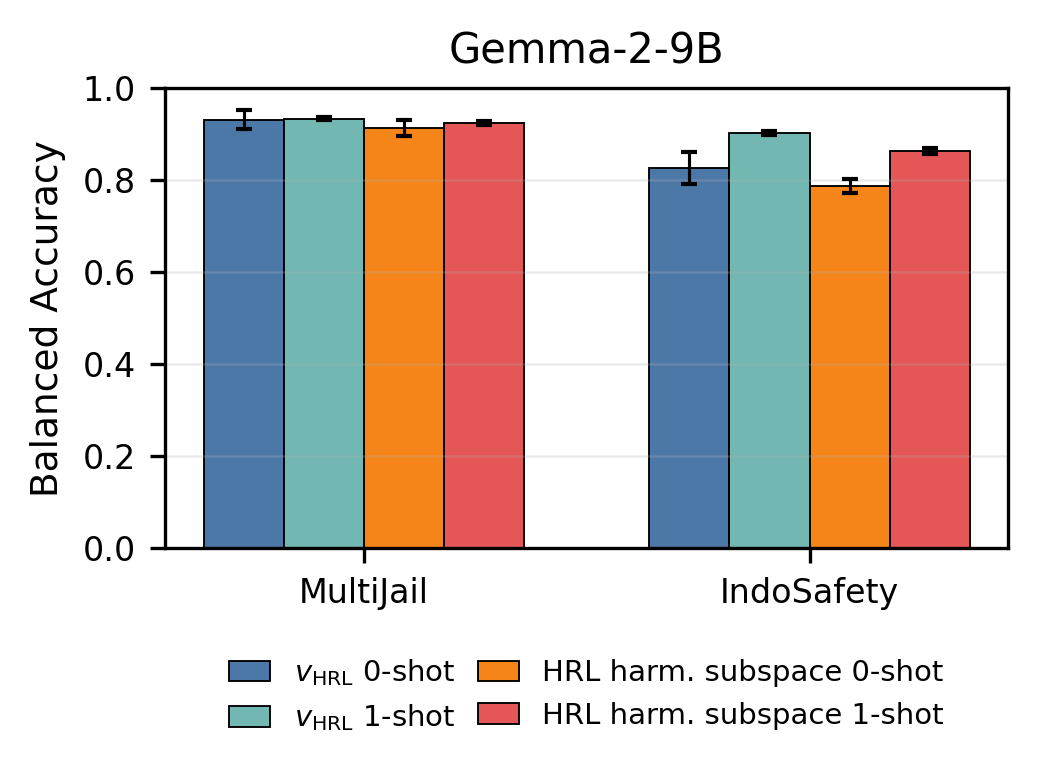}
    \caption{
    \textbf{Gemma classifiers on out-of-distribution safety benchmarks.}
    Bars report held-out balanced accuracy on MultiJail and IndoSafety for
    Gemma-2-9B, averaged within each dataset family. Error bars show the
    standard error across target languages and calibration seeds.
    }
    \label{fig:ood-latent-gates}
\end{figure}

\begin{figure}[!t]
    \centering
    \includegraphics[width=\columnwidth,keepaspectratio]{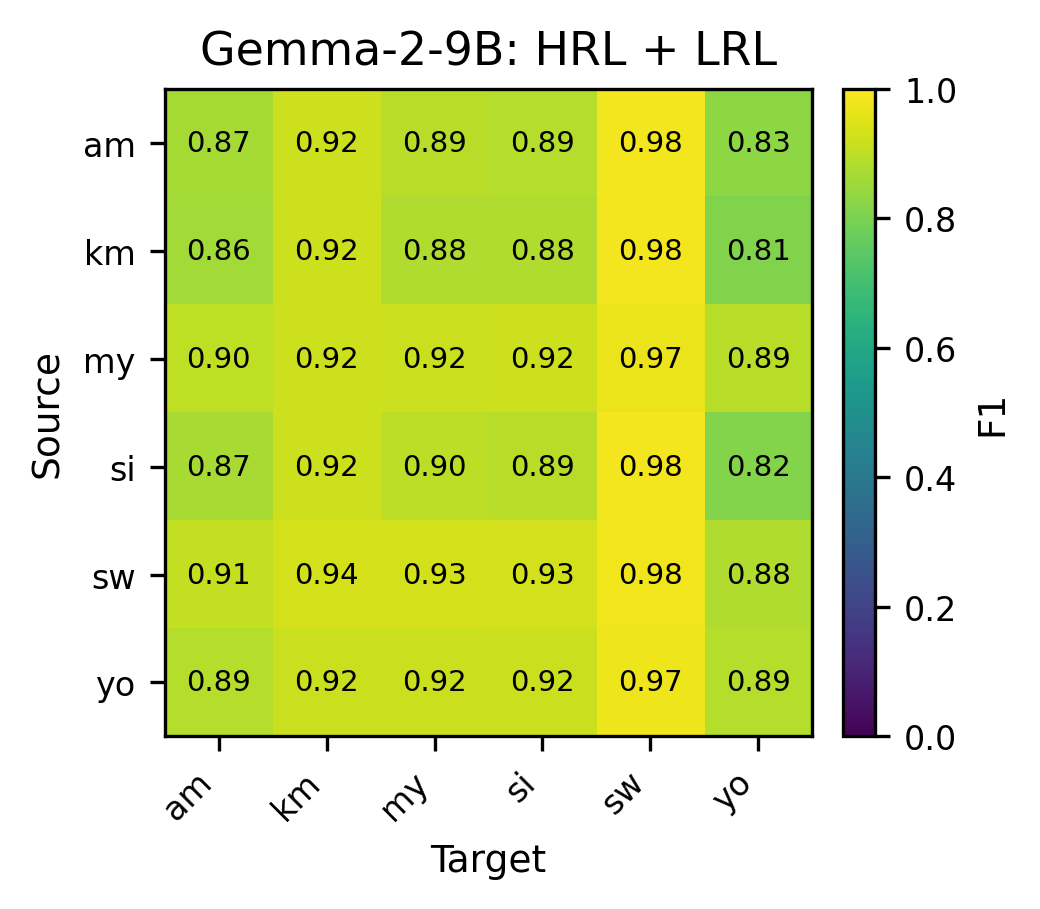}
    \caption{
    \textbf{Gemma LRL calibration often transfers across source--target pairs.}
    Cells report test F1 for an HRL subspace classifier trained with HRL data plus
    the source LRL and evaluated on the target LRL. Diagonal cells report
    held-out same-language F1.
    }
    \label{fig:crosslingual-hrl-lrl}
\end{figure}

\paragraph{The classifiers generalize beyond the calibration distribution.}
Figure~\ref{fig:ood-latent-gates} evaluates the Gemma classifiers on two
out-of-distribution safety benchmarks: MultiJail \citep{deng2023multijail} and
IndoSafety \citep{azmi-etal-2025-indosafety}. We test the transfer without calibrating
on OOD (Zero-shot) and with a single calibration example (One-shot). Transfer beyond
PolyRefuse is strong: the best Gemma classifier reaches \(0.933/0.902\) balanced accuracy on
MultiJail/IndoSafety. The full all-model results in
Figure~\ref{fig:ood-latent-gates-all-models} show the same broad pattern for
Qwen and a different one for Llama. For Qwen and Gemma, one-shot
\(v_{\mathrm{HRL}}\) is strongest; for Llama, the subspace classifier improves from
\(0.653\) to \(0.765\) on MultiJail and \(0.560\) to \(0.713\) on IndoSafety.

\paragraph{LRL calibration transfers beyond one language.}
Figure~\ref{fig:crosslingual-hrl-lrl} asks whether adding one LRL to the HRL
subspace classifier helps on other LRLs. Calibration often transfers for Gemma:
diagonal and off-diagonal means nearly match (\(0.912\) vs.\ \(0.908\)). The
all-model matrix in Figure~\ref{fig:crosslingual-hrl-lrl-all-models} shows that
Qwen has a smaller but useful pattern (\(0.843\) vs.\ \(0.792\)). The diagonal
is not always strongest; in several cases, a classifier calibrated on one LRL works
better on other LRLs than on that language's own held-out examples. This
suggests few-shot examples can expose a reusable harmfulness subspace whose
transfer depends more on target distribution than source--target language match.
Llama is the caveat, with a larger diagonal--off-diagonal gap (\(0.804\) vs.\
\(0.628\)) and more uneven transfer.

\begin{table*}[t]
\centering
\tiny
\setlength{\tabcolsep}{1.4pt}
\renewcommand{\arraystretch}{0.92}
\resizebox{\textwidth}{!}{%
\begin{tabular*}{\textwidth}{@{\extracolsep{\fill}}l*{6}{ccc}@{}}
\toprule
Model / Method & \multicolumn{3}{c}{sw} & \multicolumn{3}{c}{am} & \multicolumn{3}{c}{my} & \multicolumn{3}{c}{km} & \multicolumn{3}{c}{si} & \multicolumn{3}{c}{yo} \\
\cmidrule(lr){2-4} \cmidrule(lr){5-7} \cmidrule(lr){8-10} \cmidrule(lr){11-13} \cmidrule(lr){14-16} \cmidrule(lr){17-19}
 & Harm.$\uparrow$ & Harml.$\downarrow$ & $\Delta\uparrow$ & Harm.$\uparrow$ & Harml.$\downarrow$ & $\Delta\uparrow$ & Harm.$\uparrow$ & Harml.$\downarrow$ & $\Delta\uparrow$ & Harm.$\uparrow$ & Harml.$\downarrow$ & $\Delta\uparrow$ & Harm.$\uparrow$ & Harml.$\downarrow$ & $\Delta\uparrow$ & Harm.$\uparrow$ & Harml.$\downarrow$ & $\Delta\uparrow$ \\
\midrule
Qwen2.5-7B-Instruct & 22.9 & \underline{9.0} & 13.9 & 4.4 & \textbf{5.4} & -1.0 & 29.5 & \underline{8.4} & 21.1 & 31.1 & \underline{7.0} & 24.1 & 41.4 & \textbf{11.6} & 29.8 & 43.5 & \underline{39.8} & 3.7 \\
\quad + Ours (HRL) & \textbf{56.6} & 10.6 & \underline{46.0} & 23.3 & 6.2 & \underline{17.1} & 46.0 & \textbf{7.8} & \underline{38.2} & 55.6 & 8.6 & \underline{47.0} & 61.2 & 16.8 & \underline{44.4} & \textbf{87.6} & 48.4 & \textbf{39.2} \\
\quad + Ours (HRL + 32 LRL) & \textbf{56.6} & 10.4 & \textbf{46.2} & \underline{32.2} & 10.4 & \textbf{21.8} & \underline{51.0} & 10.8 & \textbf{40.2} & \underline{74.3} & 15.6 & \textbf{58.7} & \textbf{74.7} & 25.6 & \textbf{49.1} & \underline{87.2} & 48.6 & \underline{38.6} \\
\quad + CAST (Original) & 29.2 & 12.8 & 16.4 & 10.3 & 10.2 & 0.1 & 37.1 & 10.4 & 26.7 & 43.4 & 13.0 & 30.4 & 48.4 & 17.4 & 31.0 & 46.9 & \textbf{36.2} & 10.7 \\
\quad + CAST (HRL + 32 LRL) & 21.0 & \textbf{7.0} & 14.0 & 6.1 & \underline{6.0} & 0.1 & 30.9 & \underline{8.4} & 22.5 & 29.2 & \textbf{5.8} & 23.4 & 36.4 & \underline{13.4} & 23.0 & 50.0 & 40.8 & 9.2 \\
\quad + AdaSteer (Original) & \underline{43.5} & 25.8 & 17.7 & \textbf{71.9} & 81.6 & -9.7 & \textbf{82.0} & 66.2 & 15.8 & \textbf{79.9} & 53.0 & 26.9 & \underline{73.3} & 53.8 & 19.5 & 79.2 & 77.2 & 2.0 \\
\quad + AdaSteer (HRL + 32 LRL) & 26.7 & 16.4 & 10.3 & 9.3 & 20.6 & -11.3 & 38.1 & 18.6 & 19.5 & 40.7 & 13.2 & 27.5 & 45.6 & 20.6 & 25.0 & 44.8 & 47.8 & -3.0 \\
\midrule
Gemma-2-9B & 86.2 & \textbf{1.0} & 85.2 & 72.7 & \underline{14.4} & 58.3 & 54.9 & \textbf{3.8} & 51.1 & 74.3 & \textbf{5.4} & 68.9 & 67.0 & \underline{5.8} & 61.2 & 66.6 & 30.4 & 36.2 \\
\quad + Ours (HRL) & \underline{95.5} & 3.8 & \underline{91.7} & \underline{84.3} & \textbf{12.6} & \underline{71.7} & \underline{81.6} & 4.4 & \underline{77.2} & \underline{86.2} & 6.4 & \underline{79.8} & \underline{81.8} & \textbf{5.2} & \underline{76.6} & \underline{76.2} & \textbf{19.6} & \underline{56.6} \\
\quad + Ours (HRL + 32 LRL) & \textbf{96.7} & \underline{3.6} & \textbf{93.1} & \textbf{91.6} & 16.4 & \textbf{75.2} & \textbf{85.1} & 5.8 & \textbf{79.3} & \textbf{90.9} & 8.2 & \textbf{82.7} & \textbf{90.2} & 8.0 & \textbf{82.2} & \textbf{90.4} & \underline{25.6} & \textbf{64.8} \\
\quad + CAST (Original) & 87.6 & \textbf{1.0} & 86.6 & 77.6 & 18.0 & 59.6 & 55.6 & 4.4 & 51.2 & 79.2 & \underline{6.0} & 73.2 & 75.2 & 8.4 & 66.8 & 73.3 & 33.6 & 39.7 \\
\quad + CAST (HRL + 32 LRL) & 88.3 & \textbf{1.0} & 87.3 & 74.3 & 15.8 & 58.5 & 55.4 & \underline{4.2} & 51.2 & 81.1 & \underline{6.0} & 75.1 & 72.2 & 8.6 & 63.6 & 73.6 & 31.4 & 42.2 \\
\quad + AdaSteer (Original) & 86.4 & 4.0 & 82.4 & 70.5 & 29.6 & 40.9 & 50.7 & 14.6 & 36.1 & 74.0 & 17.0 & 57.0 & 57.3 & 16.8 & 40.5 & 70.5 & 39.4 & 31.1 \\
\quad + AdaSteer (HRL + 32 LRL) & 73.3 & 9.4 & 63.9 & 40.0 & 72.4 & -32.4 & 48.1 & 38.4 & 9.7 & 63.3 & 49.6 & 13.7 & 47.0 & 45.6 & 1.4 & 62.4 & 72.0 & -9.6 \\
\midrule
Llama-3.1-8B-Instruct & 38.6 & \textbf{0.8} & 37.8 & 9.8 & \underline{7.0} & 2.8 & 40.2 & 8.0 & 32.2 & 44.8 & 5.2 & 39.6 & 40.9 & \underline{4.8} & \underline{36.1} & 20.8 & 9.8 & 11.0 \\
\quad + Ours (HRL) & 29.0 & \underline{1.0} & 28.0 & 8.9 & \textbf{6.0} & 2.9 & 33.4 & \underline{7.6} & 25.8 & 25.3 & \textbf{3.6} & 21.7 & 31.3 & \textbf{4.6} & 26.7 & 17.7 & \textbf{8.6} & 9.1 \\
\quad + Ours (HRL + 32 LRL) & \underline{53.7} & 1.4 & \textbf{52.3} & 32.3 & 8.4 & \textbf{23.9} & 46.7 & \textbf{7.2} & \textbf{39.5} & 58.2 & \underline{4.8} & \textbf{53.4} & 64.2 & 8.0 & \textbf{56.2} & 32.7 & \underline{9.2} & \textbf{23.5} \\
\quad + CAST (Original) & 50.0 & 37.6 & 12.4 & \underline{58.9} & 53.0 & 5.9 & \underline{77.6} & 57.4 & 20.2 & \underline{70.6} & 58.0 & 12.6 & \textbf{88.5} & 80.6 & 7.9 & \underline{76.4} & 72.4 & 4.0 \\
\quad + CAST (HRL + 32 LRL) & 45.1 & 10.4 & 34.7 & 26.4 & 18.8 & 7.6 & 54.5 & 26.4 & 28.1 & 53.1 & 23.6 & 29.5 & 81.1 & 57.2 & 23.9 & 54.7 & 43.8 & 10.9 \\
\quad + AdaSteer (Original) & 46.2 & 5.0 & \underline{41.2} & 16.8 & 15.0 & 1.8 & 51.6 & 17.4 & \underline{34.2} & 55.8 & 13.6 & \underline{42.2} & 58.2 & 22.2 & 36.0 & 28.3 & 15.4 & \underline{12.9} \\
\quad + AdaSteer (HRL + 32 LRL) & \textbf{83.4} & 79.2 & 4.2 & \textbf{93.2} & 84.2 & \underline{9.0} & \textbf{91.3} & 92.2 & -0.9 & \textbf{87.2} & 87.8 & -0.6 & \underline{86.0} & 92.2 & -6.2 & \textbf{91.8} & 89.6 & 2.2 \\
\bottomrule
\end{tabular*}
}
\caption{Low-resource test refusal rates by steering method: harmful-refusal, harmless-refusal, and $\Delta$ (harmful minus harmless) percentages. Ours (HRL) uses only the HRL-fitted rank-10 conditional $v_{\mathrm{HRL}}$ subspace; Ours (HRL + 32 LRL) adds 32 harmful and 32 harmless target-language calibration examples. CAST (Original) and AdaSteer (Original) use each method's own released training data. Bold marks the best value and underline marks the second-best value within each model, language, and metric.}
\label{tab:lrl-method-comparison}
\vspace{-1.2em}
\end{table*}

\subsection{Using the Calibrated Classifier for Conditional Steering}
\label{sec:routing-refusal}

We combine the preceding analyses in a training-free steering method, evaluated
on the six PolyRefuse LRLs where the refusal gap is largest. Baselines\footnote{see Appendix \ref{app:baseline-details} for more details} are state-of-the-art latent steering methods, CAST and AdaSteer.

\begin{algorithm}
\small
\caption{Conditional HRL refusal intervention}
\label{alg:conditional-vhrl-intervention}
\begin{algorithmic}[1]
\Require Activation \(h_k(x)\), classifier logit \(g(x)\), threshold \(\tau\), Unit direction \(\hat v_{\mathrm{HRL}}\)
\Statex \(S_{\mathrm{HRL},c} \gets \{h_k(x)^\top\hat v_{\mathrm{HRL}}: x\in\mathcal{D}^{\mathrm{train}}_{\mathrm{HRL},c}\}\), \(c\in\{H,N\}\)
\Statex \(\lambda \gets Q_{0.75}(S_{\mathrm{HRL},H}) - Q_{0.25}(S_{\mathrm{HRL},N})\) \Comment{interquartile score range}
\If{\(g(x) > \tau\)}
    \State \(h_k(x) \gets h_k(x) + \lambda\,\hat v_{\mathrm{HRL}}\)
\Else
    \State \(h_k(x) \gets h_k(x) - \hat v_{\mathrm{HRL}}{\hat v_{\mathrm{HRL}}}^{T} h_k(x)\)
\EndIf
\end{algorithmic}
\end{algorithm}

\paragraph{Conditional steering.}
Following CAST \cite{lee2025cast}, the classifier chooses between adding and removing
the HRL harmfulness direction at the extraction layer. We add the direction to
prompts classified as harmful and remove it from prompts classified as harmless.
Algorithm~\ref{alg:conditional-vhrl-intervention} gives
the intervention, where $g$ is the subspace classifier from
Algorithm~\ref{alg:hrl-subspace-gate} and $\tau$ is calibrated on the same data.
We set $\lambda$ as the gap between HRL harmful upper-quartile and harmless
lower-quartile projection scores. This moves low-resource harmful activations toward the
high-resource harmful score regime without tuning a language-specific steering strength.
The classifier computes a rank-10 projection once per prompt, and generation adds or
removes one direction at one layer, an $O(d)$ operation in the residual width.
This requires no additional model forward pass, external guard model, or weight
update and has the same asymptotic intervention overhead as CAST and AdaSteer.

\paragraph{Experimental setup.}
We train or calibrate each method with matched data: \textbf{Original} uses each
prior method's released training data; \textbf{HRL} uses 200 PolyRefuse HRL
examples per language and class; \textbf{HRL + 32 LRL} adds 32 target-language
LRL examples per class.

\begin{table}[t]
\centering
\scriptsize
\setlength{\tabcolsep}{2.0pt}
\renewcommand{\arraystretch}{1.0}
\resizebox{\columnwidth}{!}{%
\begin{tabular}{lccccc}
\toprule
Model & sw & am & si & yo & Avg. $\Delta$ \\
\midrule
Qwen2.5-7B & 26.9$\rightarrow$27.8 & 26.2$\rightarrow$26.2 & 28.0$\rightarrow$28.2 & 27.9$\rightarrow$28.4 & +0.4 \\
Gemma-2-9B & 28.9$\rightarrow$30.1 & 23.1$\rightarrow$23.2 & 27.7$\rightarrow$28.9 & 23.0$\rightarrow$23.0 & +0.6 \\
Llama-3.1-8B & 34.5$\rightarrow$34.4 & 30.3$\rightarrow$30.8 & 33.4$\rightarrow$33.1 & 25.3$\rightarrow$25.3 & +0.0 \\
\bottomrule
\end{tabular}%
}
\caption{\textbf{Utility is preserved after steering.} Global-MMLU accuracy (\%) before and after \textit{Ours (HRL + 32 LRL)} conditional steering.}
\label{tab:mmmlu-conditional-utility}
\end{table}

\paragraph{HRL-only calibration works for Qwen and Gemma.}
Table~\ref{tab:lrl-method-comparison} shows that an HRL harmfulness subspace can
support conditional steering without target-language calibration. With HRL data only (Zero-shot),
our method raises mean \(\Delta\) from \(15.3\) to \(38.6\) on Qwen and \(60.1\)
to \(75.6\) on Gemma. The gain is selective: averaged across model--language
pairs, the HRL-only classifier reaches \(54.5\%\) harmful refusal with \(10.1\%\)
harmless refusal, while CAST (Original) and AdaSteer (Original) reach higher
harmful refusal but much higher harmless refusal (\(29.5\%\) and \(31.5\%\)).
Llama is the exception, where the HRL-only classifier underperforms the base model,
matching the earlier finding that HRL harmfulness scores are not automatically
calibrated for every model and language. Its mean LRL selectivity falls from
26.6 in the base model to 19.0 with the HRL-only classifier. HRL-only calibration
often transfers for Qwen and Gemma, whereas Llama requires target-language
calibration in these experiments.

\paragraph{A small LRL budget closes most gaps.}
Adding 32 harmful and 32 harmless target-language examples improves \(\Delta\)
in 17 of 18 model--language pairs. The gain is largest for Llama, where mean
\(\Delta\) moves from \(19.0\) with the HRL-only classifier to \(41.5\) with HRL + 32
LRL; Qwen and Gemma improve from \(38.6\) to \(42.4\) and \(75.6\) to \(79.6\).
Target-language examples therefore calibrate the classifier without learning a new
harmfulness direction. For Llama, the resulting 41.5 selectivity exceeds the
base model in all six focal LRLs.

\paragraph{Our method is most selective.}
HRL + 32 LRL achieves the best $\Delta$ in 17 of 18 pairs, reaching $54.5$ mean $\Delta$, substantially above the strongest adapted baseline, CAST (HRL + 32 LRL) at $33.6$. CAST keeps harmless refusal low ($18.3\%$) but only reaches $51.9\%$ harmful refusal; our method lowers harmless refusal to $12.7\%$ while raising harmful refusal to $67.2\%$. Utility is preserved on multilingual MMLU, with minor gains for Qwen and Gemma (Table~\ref{tab:mmmlu-conditional-utility}). AdaSteer under the same HRL + 32 LRL setup over-refuses benign prompts, averaging $52.8\%$ harmless refusal and only $6.8$ mean $\Delta$.

\paragraph{Our method adds little inference overhead.}
In a generation benchmark with fixed 128-token outputs, our method adds
only $0.7\%$, $2.2\%$, and $1.8\%$ mean latency on Gemma, Llama, and Qwen,
respectively. The corresponding overhead is $3.4\%$, $12.3\%$, and $13.6\%$
for adapted CAST and $8.7\%$, $10.1\%$, and $9.7\%$ for adapted AdaSteer.
Thus, the rank-10 gate's single prompt-level projection produces consistently
lower empirical overhead than previous baselines (Table~\ref{tab:intervention-overhead}).

\subsection{Analyzing Classifier Errors and Post-Intervention Non-Refusals}
\label{sec:three-stage-diagnosis}

We use our learned classifier and intervention to divide the remaining harmful
non-refusals into two groups. Under HRL + 32 LRL, a \emph{classifier error}
occurs when our classifier labels a harmful prompt as harmless. A
\emph{post-intervention non-refusal} occurs when the classifier labels it as
harmful and the method adds the direction, but the model still does not refuse.

\begin{table}[t]
\centering
\small
\setlength{\tabcolsep}{5pt}
\begin{tabular}{lrr}
\toprule
Model & Classifier error & Post-intervention non-refusal \\
\midrule
Qwen  & 32.2\% & 67.8\% \\
Gemma & 93.7\% &  6.3\% \\
Llama & 49.7\% & 50.3\% \\
\bottomrule
\end{tabular}
\caption{\textbf{Sources of remaining harmful non-refusals under HRL + 32 LRL.}
Each row partitions unrefused harmful prompts according to whether our classifier
labels the prompt harmless or labels it harmful and direction addition still does
not produce refusal.}
\label{tab:residual-error-summary}
\end{table}

Table~\ref{tab:residual-error-summary} partitions the remaining harmful
non-refusals. Gemma's residual failures are primarily classifier errors, Qwen's
are primarily post-intervention non-refusals, and Llama exhibits both. Threshold calibration
would address most Gemma failures, but not most Qwen failures. The full
classifier-prediction $\times$ final-refusal cross-tabulation and per-language shares can be seen in
Appendix~\ref{app:residual-failure-decomposition}.

\paragraph{Boundary cases.}
Unrefused harmful prompts lie closer to the classifier boundary than prompts corrected
by the intervention for all three models. Newly over-refused harmless prompts
are likewise closer to the boundary than prompts that remain helpful. This
supports calibration as one source of both under- and over-refusal, while the
frequency of post-intervention non-refusals shows that boundary ambiguity does not explain every failure
(Appendix~\ref{app:gate-margin-analysis}).

\section{Conclusion and Discussion}

\paragraph{Conclusion.} In our studied settings, complete absence of harmfulness information cannot explain many low-resource safety failures. Harmful and harmless prompts remain separable in the HRL direction, although their scores can weaken or shift. Adding or ablating that direction changes refusal. Remaining failures include errors by our learned classifier and non-refusals after direction addition. The classifier-guided intervention improves selective refusal across most model--language pairs while preserving benign instruction following better than adapted baselines.

\paragraph{Implications.}
Behavioral refusal rates alone conflate two distinct failure modes: a model that lacks a harmfulness representation in a given language, and a model that has one but fails to act on it. Our results show these can come apart, and that current low-resource refusal gaps in the studied models reflect the second case more than the first. The two failures call for different fixes: missing representations call for better representation learning, while weakly routed ones can be repaired with small target-language calibration. The practical path is a diagnostic-then-fix workflow: test whether harmfulness is represented before deciding how to repair its use.



\section*{Limitations}

\paragraph{Scope.}
We study three open dense instruction-tuned 7B--9B models and 23 PolyRefuse
languages; frontier, MoE/sparse models and other languages may use different
mechanisms, and Common Crawl share is a coarse resource proxy.
\paragraph{Evaluation and access.}
We score greedy completions with GPT-4o-mini, so automatic judgments can miss
multilingual nuance; interventions require activation access and are diagnostics,
not closed-model safeguards.
\paragraph{Training coverage.}
Gated steering calibrates when to refuse; it does not replace multilingual safety
training or guarantee refusal wording, harmful-category coverage, or
adversarial-prompt robustness.

\bibliography{references,anthology-1}

\clearpage

\appendix

\section{Dataset Details}
\label{app:polyrefuse-details}
\label{app:dataset-details}

\subsection{PolyRefuse}
\label{app:dataset-polyrefuse}
PolyRefuse is the main safety dataset in our experiments. Its harmful prompts
come from AdvBench \citep{zou2023universal}, MaliciousInstruct
\citep{huang2024catastrophic}, and TDC2023 \citep{tdc2023}; its harmless prompts
come from Alpaca \citep{taori2023stanfordalpaca}. The original release
uses Google Translate to render these English prompts in 13 non-English languages
\citep{wang2025refusaluniversal}. We keep those languages and add Greek, Hebrew,
Hindi, Persian, Swahili, Amharic, Burmese, Khmer, and Sinhala using Google Translate.
We measure translation quality using SBERT \cite{reimers-gurevych-2019-sentence}
and find high semantic similarity after back-translation
(Table~\ref{tab:polyrefuse-test-sbert}).

We evaluate 23 languages. English, German, French, Spanish, Italian, Dutch,
Polish, Russian, Chinese, and Japanese form the HRL tier; Arabic, Korean, Thai,
Greek, Hebrew, Hindi, and Persian form the MRL tier; Swahili, Amharic, Burmese,
Khmer, Sinhala, and Yoruba form the LRL tier. Tiers follow Common Crawl share:
HRL above 1\%, MRL between 0.1\% and 1\%, and LRL below 0.1\%. Each language has
260/39/572 harmful train/validation/test prompts and 200/200/500 harmless
train/validation/test prompts. Table~\ref{tab:polyrefuse-language-families}
summarizes the language-family coverage of the extension.

\definecolor{tblMidYellow}{HTML}{FFF2CC}
\definecolor{tblWeakOrange}{HTML}{FCE5CD}
\definecolor{tblBadRed}{HTML}{F4CCCC}
\newcommand{\tblbest}[1]{#1}
\newcommand{\tblgood}[1]{#1}
\newcommand{\tblok}[1]{\cellcolor{tblMidYellow}#1}
\newcommand{\tblweak}[1]{\cellcolor{tblWeakOrange}#1}
\newcommand{\tblbad}[1]{\cellcolor{tblBadRed}#1}

\begin{table*}[!tp]
\centering
\small
\setlength{\tabcolsep}{4.8pt}
\renewcommand{\arraystretch}{1.08}
\begin{tabular*}{\textwidth}{@{\extracolsep{\fill}}llcccccc@{}}
\toprule
& & \multicolumn{2}{c}{Qwen2.5-7B-Instruct} & \multicolumn{2}{c}{Gemma-2-9B-it} & \multicolumn{2}{c}{Llama-3.1-8B-Instruct} \\
\cmidrule(lr){3-4}\cmidrule(lr){5-6}\cmidrule(l){7-8}
Tier & Language
& \shortstack{Harmful$\uparrow$}
& \shortstack{Harmless$\downarrow$}
& \shortstack{Harmful$\uparrow$}
& \shortstack{Harmless$\downarrow$}
& \shortstack{Harmful$\uparrow$}
& \shortstack{Harmless$\downarrow$} \\
\midrule
\textsc{High} & English (en) & \tblgood{89.5} & \tblbest{0.4} & \tblbest{97.9} & \tblgood{2.0} & \tblbest{92.7} & \tblbest{0.4} \\
 & German (de) & \tblgood{87.2} & \tblbest{1.0} & \tblbest{95.5} & \tblgood{1.2} & \tblgood{81.3} & \tblbest{0.2} \\
 & French (fr) & \tblgood{85.1} & \tblbest{1.0} & \tblbest{95.3} & \tblgood{1.6} & \tblbest{90.7} & \tblbest{0.6} \\
 & Spanish (es) & \tblgood{85.7} & \tblgood{1.2} & \tblbest{94.2} & \tblgood{1.4} & \tblgood{87.9} & \tblbest{0.4} \\
 & Italian (it) & \tblgood{87.2} & \tblgood{1.2} & \tblbest{95.6} & \tblgood{1.4} & \tblgood{87.4} & \tblbest{0.6} \\
 & Dutch (nl) & \tblgood{85.3} & \tblbest{1.0} & \tblbest{95.3} & \tblbest{1.0} & \tblgood{88.3} & \tblgood{1.2} \\
 & Polish (pl) & \tblgood{87.6} & \tblbest{0.8} & \tblbest{94.9} & \tblgood{1.8} & \tblgood{80.6} & \tblbest{0.4} \\
 & Russian (ru) & \tblgood{86.4} & \tblbest{0.8} & \tblbest{94.2} & \tblgood{2.0} & \tblgood{86.9} & \tblbest{0.8} \\
 & Chinese (zh) & \tblgood{81.3} & \tblbest{0.4} & \tblbest{92.1} & \tblgood{2.0} & \tblgood{82.0} & \tblbest{0.6} \\
 & Japanese (ja) & \tblgood{80.8} & \tblbest{0.8} & \tblgood{89.7} & \tblgood{2.4} & \tblok{57.0} & \tblbest{0.6} \\
\rowcolor{gray!8}
 & \textit{Average} & \tblgood{\textit{85.6}} & \tblbest{\textit{0.9}} & \tblbest{\textit{94.5}} & \tblgood{\textit{1.7}} & \tblgood{\textit{83.5}} & \tblbest{\textit{0.6}} \\
\midrule
\textsc{Medium} & Arabic (ar) & \tblgood{82.5} & \tblbest{1.0} & \tblbest{93.5} & \tblgood{2.8} & \tblgood{79.7} & \tblgood{1.8} \\
 & Korean (ko) & \tblgood{77.3} & \tblgood{1.4} & \tblgood{88.3} & \tblgood{1.6} & \tblok{50.5} & \tblbest{0.0} \\
 & Thai (th) & \tblgood{81.5} & \tblgood{1.2} & \tblbest{92.8} & \tblgood{2.4} & \tblok{71.3} & \tblgood{1.6} \\
 & Greek (el) & \tblgood{76.6} & \tblgood{5.2} & \tblbest{94.4} & \tblbest{0.4} & \tblgood{75.2} & \tblbest{0.4} \\
 & Hebrew (he) & \tblok{69.6} & \tblgood{3.4} & \tblbest{93.5} & \tblbest{0.8} & \tblgood{76.6} & \tblgood{1.8} \\
 & Hindi (hi) & \tblok{66.3} & \tblgood{3.4} & \tblbest{92.7} & \tblgood{1.4} & \tblok{67.0} & \tblgood{1.2} \\
 & Persian (fa) & \tblok{67.0} & \tblgood{2.0} & \tblbest{92.3} & \tblgood{1.6} & \tblok{68.0} & \tblbest{0.8} \\
\rowcolor{gray!8}
 & \textit{Average} & \tblok{\textit{74.4}} & \tblgood{\textit{2.5}} & \tblbest{\textit{92.5}} & \tblgood{\textit{1.6}} & \tblok{\textit{69.8}} & \tblgood{\textit{1.1}} \\
\midrule
\textsc{Low} & Swahili (sw) & \tblbad{22.9} & \tblgood{9.0} & \tblgood{86.2} & \tblbest{1.0} & \tblweak{38.6} & \tblbest{0.8} \\
 & Amharic (am) & \tblbad{4.4} & \tblgood{5.4} & \tblok{72.7} & \tblweak{14.4} & \tblbad{9.8} & \tblgood{7.0} \\
 & Burmese (my) & \tblweak{29.5} & \tblgood{8.4} & \tblok{54.9} & \tblgood{3.8} & \tblweak{40.2} & \tblgood{8.0} \\
 & Khmer (km) & \tblweak{31.1} & \tblgood{7.0} & \tblok{74.3} & \tblgood{5.4} & \tblweak{44.8} & \tblgood{5.2} \\
 & Sinhala (si) & \tblweak{41.4} & \tblok{11.6} & \tblok{67.0} & \tblgood{5.8} & \tblweak{40.9} & \tblgood{4.8} \\
 & Yoruba (yo) & \tblweak{43.5} & \tblbad{39.8} & \tblok{66.6} & \tblbad{30.4} & \tblbad{20.8} & \tblgood{9.8} \\
\rowcolor{gray!8}
 & \textit{Average} & \tblweak{\textit{28.8}} & \tblweak{\textit{13.5}} & \tblok{\textit{70.3}} & \tblok{\textit{10.1}} & \tblweak{\textit{32.5}} & \tblgood{\textit{5.9}} \\
\bottomrule

\end{tabular*}

\caption{
Refusal rates (\%) by language and resource tier.
}
\label{tab:main-tier-results}

\end{table*}

\begin{table}[H]
\centering
\scriptsize
\setlength{\tabcolsep}{3pt}
\renewcommand{\arraystretch}{1.05}
\begin{tabular}{p{0.25\columnwidth}p{0.56\columnwidth}r}
\toprule
Language family & Languages & Count \\
\midrule
Indo-European & English, German, French, Spanish, Italian, Dutch, Polish,
Russian, Greek, Hindi, Persian, Sinhala & 12 \\
Afro-Asiatic & Arabic, Hebrew, Amharic & 3 \\
Sino-Tibetan & Chinese, Burmese & 2 \\
Niger-Congo & Swahili, Yoruba & 2 \\
Japonic & Japanese & 1 \\
Koreanic & Korean & 1 \\
Kra-Dai & Thai & 1 \\
Austroasiatic & Khmer & 1 \\
\bottomrule
\end{tabular}
\caption{
Language-family coverage in the 23-language PolyRefuse extension.
}
\label{tab:polyrefuse-language-families}
\end{table}

\subsection{Translation and Yoruba Sensitivity}
\label{app:translation-sensitivity}

The SBERT back-translation analysis in
Table~\ref{tab:polyrefuse-test-sbert} tests approximate semantic preservation.

\begin{table*}
\centering
\small
\setlength{\tabcolsep}{10pt}

\begin{tabular}{lccccccccc}
\toprule
\multicolumn{10}{c}{\textbf{High-resource languages}} \\
\midrule
Subset & de & fr & es & it & nl & pl & ru & zh & ja \\
\midrule
Harmful & 95.21 & 96.19 & 96.11 & 96.39 & 96.38 & 93.71 & 94.83 & 93.55 & 93.22 \\
Harmless & 96.51 & 97.53 & 97.68 & 97.59 & 97.73 & 93.83 & 95.96 & 95.45 & 95.51 \\
\bottomrule
\end{tabular}

\vspace{0.75em}

\begin{tabular}{lccccccc}
\toprule
\multicolumn{8}{c}{\textbf{Mid-resource languages}} \\
\midrule
Subset & ar & ko & th & el & he & hi & fa \\
\midrule
Harmful & 95.13 & 92.73 & 90.31 & 95.98 & 95.97 & 95.91 & 94.54 \\
Harmless & 96.05 & 94.86 & 93.25 & 96.56 & 96.91 & 96.46 & 94.71 \\
\bottomrule
\end{tabular}

\vspace{0.75em}

\begin{tabular}{lcccccc}
\toprule
\multicolumn{7}{c}{\textbf{Low-resource languages}} \\
\midrule
Subset & sw & am & my & km & si & yo \\
\midrule
Harmful & 92.41 & 92.14 & 91.76 & 88.05 & 94.47 & 91.81 \\
Harmless & 94.15 & 94.40 & 94.29 & 91.52 & 96.17 & 93.15 \\
\bottomrule
\end{tabular}

\caption{SBERT back-translation similarity on PolyRefuse across train, validation, and test splits. Target-language prompts are back-translated to English and compared with the original English prompts; scores are cosine similarities scaled to 0--100.}
\label{tab:polyrefuse-test-sbert}
\end{table*}

It does not measure whether translations are natural, whether harmfulness has
the same cultural interpretation, or whether resource tier causes the observed
refusal gap. Our PolyRefuse results therefore support a within-benchmark
representation--behavior mismatch on parallel prompts.

Yoruba has substantially higher harmless refusal than the other focal LRLs.
Table~\ref{tab:yoruba-sensitivity} repeats the aggregate behavioral analysis
without it. Base harmful refusal remains 43.9\%; under HRL + 32 LRL, harmful
refusal changes from 67.2\% to 66.6\%, while selectivity rises from 54.5 to
56.9. Yoruba mainly affects harmless refusal and is not representative of the
other five LRLs on this measure.

\begin{table}[H]
\centering
\scriptsize
\setlength{\tabcolsep}{3pt}
\begin{tabular}{p{0.48\columnwidth}rr}
\toprule
Macro-average across models & \shortstack{All 6\\LRLs} & \shortstack{Without\\Yoruba} \\
\midrule
Base harmful refusal & 43.9\% & 43.9\% \\
Base harmless refusal & 9.9\% & 6.5\% \\
Ours harmful refusal & 67.2\% & 66.6\% \\
Ours harmless refusal & 12.7\% & 9.6\% \\
Ours selectivity & 54.5 & 56.9 \\
\bottomrule
\end{tabular}
\caption{\textbf{Sensitivity to the Yoruba over-refusal outlier.} ``Ours'' is
HRL + 32 LRL; selectivity is harmful minus harmless refusal in percentage points.}
\label{tab:yoruba-sensitivity}
\end{table}

\subsection{Global-MMLU}
\label{app:dataset-global-mmlu}
Global-MMLU \citep{singh-etal-2025-global} is our utility benchmark. It extends
MMLU for 42 languages using professional translations, crowdsourced
translations, machine translations, human quality review, and post-edits. It
also marks questions that depend on cultural, geographic, or dialect knowledge.
We use it only for Table~\ref{tab:mmmlu-conditional-utility}.

Global-MMLU overlaps with four of our six LRLs: Amharic, Sinhala, Swahili, and
Yoruba. Burmese and Khmer are missing, so the Global-MMLU utility table covers
the four overlapping languages.

\subsection{IndoSafety}
\label{app:dataset-indosafety}
IndoSafety \citep{azmi-etal-2025-indosafety} is a culturally grounded safety
benchmark for formal Indonesian, colloquial Indonesian, Javanese, Sundanese, and
Minangkabau. The benchmark extends beyond translated English harmful prompts by
adding Indonesian-specific risks such as religious and ethnic
sensitivities, regional separatism, Pancasila misinterpretation, Indonesian
public entities, local cultural practices, and supernatural claims.

The dataset is human-verified and culturally localized. The formal Indonesian
set includes translated and manually edited safety prompts, plus
Indonesian-specific prompts written and checked by native speakers with local
cultural and safety expertise. For the colloquial and local-language extension,
the authors start with Google Translate for Javanese, Sundanese, and Minangkabau
and GPT-4o for colloquial Indonesian, then have native speakers review and edit
the outputs.

\subsection{MultiJail}
\label{app:dataset-multijail}
MultiJail \citep{deng2023multijail} provides multilingual jailbreak prompts. It
starts from 315 English unsafe prompts and adds native-speaker translations into
Chinese, Italian, Vietnamese, Arabic, Korean, Thai, Bengali, Swahili, and
Javanese. The original benchmark tests both direct multilingual prompts and
prompts paired with an English jailbreak instruction. We only make use of the
harmful multilingual prompts.

\section{Decoding Parameters}
\label{app:decoding-parameters}

Completions use each model's chat template and greedy decoding. We generate at
most 128 new tokens, disable sampling, and use no custom stop strings. The
refusal judge is run deterministically as well through OpenRouter.

\section{Baseline Method Details}
\label{app:baseline-details}

Baseline comparisons use released artifacts when available. The HRL variants
train on the ten PolyRefuse HRLs, and the HRL + 32 LRL variants add 32 harmful
and 32 harmless examples from the target LRL.

\paragraph{CAST.}
Conditional Activation Steering (CAST) \citep{lee2025cast} learns a condition
vector and a behavior vector, then steers when the condition score (computed via cosine similarity)
crosses a threshold. Our adapted version keeps this rule and estimates both directions
from PolyRefuse contrastive activations. The condition layer and threshold are
selected on held-out PolyRefuse data.

\paragraph{AdaSteer.}
AdaSteer \citep{zhao-etal-2025-adasteer} combines a Rejection Direction (RD) and
a Harmfulness Direction (HD). Its steering coefficients are prompt-dependent and
are computed from the final prompt-token residual. We use released vectors when
available; otherwise we estimate RD and HD from the same PolyRefuse budget used
by the other adapted baselines.

\paragraph{OMNIGuard / MultiGuard.}
OMNIGuard, released as MultiGuard \citep{verma-etal-2025-multiguard}, chooses a layer
with a Universality Score (U-Score) and trains a harmfulness classifier on
mean-pooled prompt representations. We compute U-Score on English-to-HRL
translation pairs and train the classifier on PolyRefuse HRLs. With nonzero LRL
budgets, we fine-tune the classifier on the same target examples used by the
other adapted baselines.

\section{Compute Budget}
\label{app:compute}

We did not record exact GPU-hours. All experiments ran on a single RTX 5090
rented through RunPod. Table~\ref{tab:model-licenses} lists the repositories and
licenses for the model weights used in the main experiments.

\begin{table*}[t]
\centering
\small
\setlength{\tabcolsep}{8pt}
\begin{tabular}{@{}lccc@{}}
\toprule
& \multicolumn{3}{c}{Latency overhead vs. vanilla generation (\%)} \\
\cmidrule(l){2-4}
Model & Ours (HRL + 32 LRL) & CAST (HRL + 32 LRL) & AdaSteer (HRL + 32 LRL) \\
\midrule
Gemma-2-9B   & \textbf{$0.69\;[-0.02,\,1.32]$} & $3.37\;[2.61,\,4.20]$   & $8.67\;[7.76,\,9.52]$ \\
Llama-3.1-8B & \textbf{$2.19\;[1.61,\,2.83]$}  & $12.31\;[11.62,\,13.12]$ & $10.09\;[9.50,\,10.70]$ \\
Qwen2.5-7B   & \textbf{$1.79\;[1.01,\,2.51]$}  & $13.59\;[12.50,\,14.59]$ & $9.72\;[8.82,\,10.72]$ \\
\bottomrule
\end{tabular}
\caption{Values are the percentage latency overhead relative to vanilla
generation with no activation intervention. We benchmark the PolyRefuse test split in six LRLs 
(Swahili, Amharic, Burmese, Khmer, Sinhala, and Yoruba), using both harmful and harmless subsets 
with 16 prompts per language--split, batch size 4, and 128 newly generated tokens per prompt.
Measurements include one warmup followed by three timed repeats of
\texttt{model.generate}; tokenization and input preparation are excluded, and
all model and intervention parameters are preloaded on a single NVIDIA GeForce
RTX 5090 in bfloat16. 95\% Confidence intervals use 2,000 paired bootstrap samples.}
\label{tab:intervention-overhead}
\end{table*}

\begin{table}[t]
\centering
\scriptsize
\setlength{\tabcolsep}{4pt}
\begin{tabular}{@{}p{0.24\columnwidth}p{0.66\columnwidth}@{}}
\toprule
Model & Repository and license \\
\midrule
Qwen2.5-7B & Qwen/Qwen2.5-7B-Instruct; Apache License 2.0 \\
Gemma-2-9B & google/gemma-2-9b-it; Gemma Terms of Use \\
Llama-3.1-8B & meta-llama/Llama-3.1-8B-Instruct; Llama 3.1 Community License \\
\bottomrule
\end{tabular}
\caption{Model repositories and licenses, verified from the corresponding model
cards and repository license files.}
\label{tab:model-licenses}
\end{table}

\section{Residual Failure Decomposition}
\label{app:residual-failure-decomposition}

We cross-tabulate our classifier's prediction and the final response for every harmful
PolyRefuse test prompt under HRL + 32 LRL. Table~\ref{tab:residual-gate-output-crosstab}
reports the four possible outcomes as shares of all harmful prompts and then
restricts attention to remaining non-refusals. Qwen--Amharic confirms that a
harmful classifier prediction need not produce refusal: the classifier labels
83.6\% of harmful prompts as harmful, yet 52.1\% of all harmful prompts are
post-intervention non-refusals. Table~\ref{tab:residual-error-by-language} gives the
corresponding decomposition for every focal model--language pair.

\begin{table*}[t]
\centering
\scriptsize
\setlength{\tabcolsep}{4pt}
\begin{tabular}{lrrrrr}
\toprule
Model & Classifier: harmful, & Classifier: harmful, & Classifier: harmless, & Classifier: harmless, & Post-intervention among \\
 & refused & not refused & refused & not refused & remaining failures \\
\midrule
Qwen & 59.2\% & 25.3\% & 3.4\% & 12.0\% & 67.8\% (869/1,281) \\
Gemma & 90.0\% & 0.6\% & 0.8\% & 8.6\% & 6.3\% (20/315) \\
Llama & 41.7\% & 26.2\% & 6.3\% & 25.9\% & 50.3\% (898/1,786) \\
Qwen--Amharic & 31.5\% & 52.1\% & 0.7\% & 15.7\% & 76.8\% (298/388) \\
\bottomrule
\end{tabular}
\caption{\textbf{Classifier prediction $\times$ final refusal for harmful PolyRefuse prompts under HRL + 32 LRL.}
The first four columns are percentages of all harmful test prompts. The final
column is restricted to harmful prompts that remain unrefused.}
\label{tab:residual-gate-output-crosstab}
\end{table*}

\begin{table}[t]
\centering
\scriptsize
\setlength{\tabcolsep}{4pt}
\begin{tabular}{lrrrrrr}
\toprule
Model & sw & am & my & km & si & yo \\
\midrule
Qwen  & 29/71 & 23/77 & 29/71 & 63/37 & 33/67 & 38/62 \\
Gemma & 58/42 & 100/0 & 89/11 & 100/0 & 95/5 & 100/0 \\
Llama & 61/39 & 54/46 & 70/30 & 53/47 & 40/60 & 26/74 \\
\bottomrule
\end{tabular}
\caption{\textbf{Residual error shares by model and language.}
Each cell reports classifier error / post-intervention non-refusal (\%) among harmful prompts
that remain unrefused. Languages are Swahili (sw), Amharic (am), Burmese (my),
Khmer (km), Sinhala (si), and Yoruba (yo).}
\label{tab:residual-error-by-language}
\end{table}

\section{Out-of-Distribution Evaluation on MultiJail and IndoSafety}
\label{app:ood-setup}
\label{app:native-evaluation}

The OOD transfer experiment keeps the classifier fixed and changes the harmful prompt
distribution. Harmful prompts come from MultiJail or IndoSafety
(Appendix~\ref{app:dataset-multijail}; Appendix~\ref{app:dataset-indosafety}).

\paragraph{Harmful prompts.}
MultiJail contributes harmful-only multilingual jailbreak prompts. IndoSafety
contributes Indonesian and Indonesian-family harmful prompts with risk-area and
harm-type metadata. We use the provided formal-Indonesian split and test-only
splits for the regional-language cells.

\paragraph{Harmless prompts.}
The OOD benchmarks do not provide matched harmless prompts, so we pair their
harmful examples with PolyRefuse harmless examples in the same language. For
languages outside the PolyRefuse harmless set, we translate the English harmless
splits. This changes the harmful side while keeping the harmless side comparable
to the main task.

\begin{figure*}[!htbp]
    \centering
    \includegraphics[width=\textwidth]{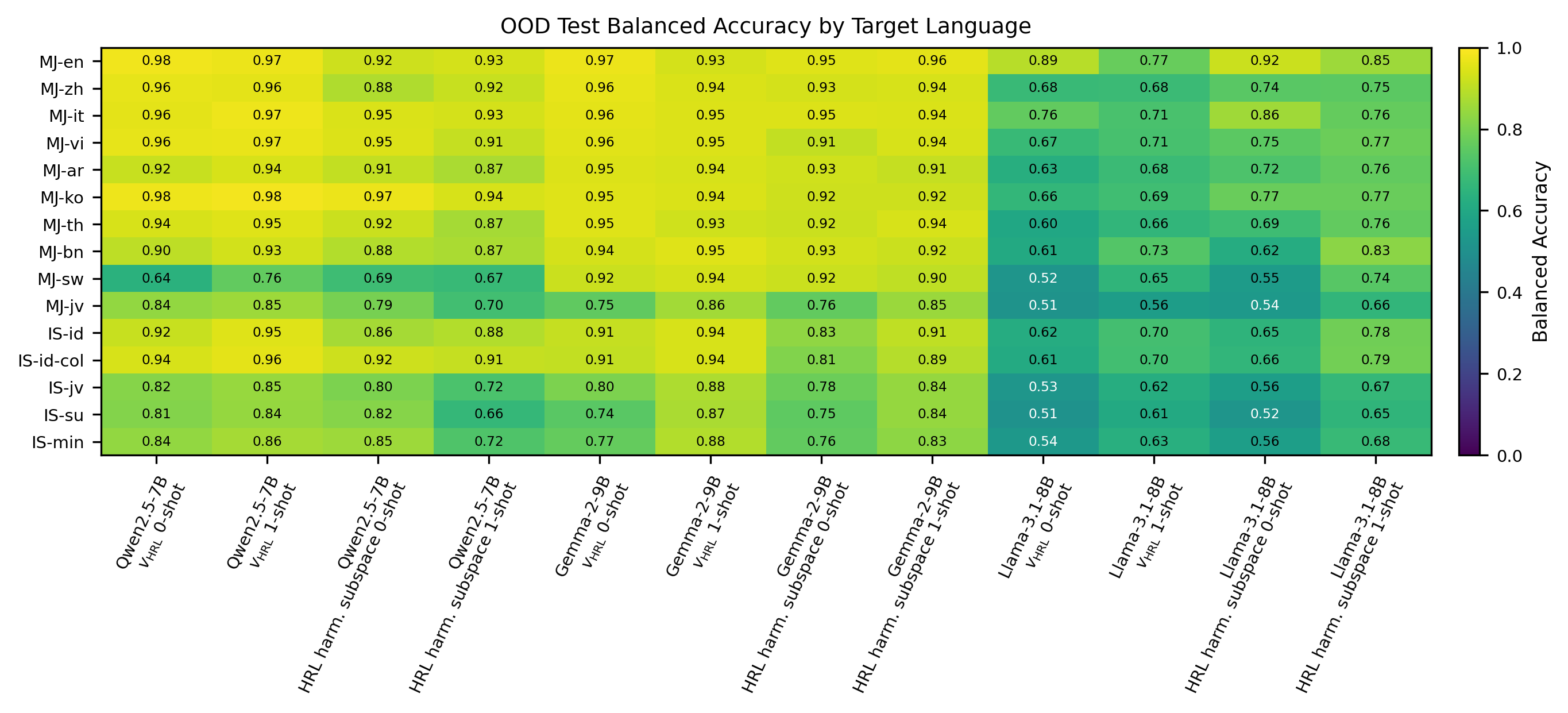}
    \caption{
    \textbf{OOD transfer by target language.}
    One-shot \(v_{\mathrm{HRL}}\) is strongest for Qwen and Gemma; Llama gains
    more from the HRL subspace classifier.
    }
    \label{fig:ood-language-breakdown}
\end{figure*}

\paragraph{Classifier evaluation.}
OOD classifiers use the same extraction layers as the main experiments. The zero-shot
\(v_{\mathrm{HRL}}\) classifier keeps the HRL threshold. The one-shot setting chooses a
balanced threshold from one harmful and one harmless target-language example.
The HRL subspace classifier uses the rank-10 harmfulness scores in
Algorithm~\ref{alg:hrl-subspace-gate}. Figure~\ref{fig:ood-language-breakdown}
shows the per-language OOD transfer results. IndoSafety category plots fix the
harmless side and vary only the harmful risk area or harm type.

MultiJail uses native-speaker translations of harmful prompts, including
Swahili. The best Gemma classifier reaches 0.933 balanced accuracy. IndoSafety
contains locally written or native-edited harms; the best one-shot classifiers reach
0.893, 0.902, and 0.713 balanced accuracy for Qwen, Gemma, and Llama,
respectively. These benchmarks provide evidence beyond Google-translated
PolyRefuse, although they do not provide community-authored data for all six
focal LRLs.

\paragraph{IndoSafety intervention outcomes.}
We evaluate the unmodified model, classifier-guided intervention, and intervention on every
harmful prompt across formal Indonesian, colloquial Indonesian, Javanese,
Sundanese, and Minangkabau, averaging the classifier-guided results over 10 seeds. The classifier-guided
method improves refusal by 44--74 percentage points over the unmodified models
and comes within 2--4 points of always intervening
(Table~\ref{tab:indosafety-intervention}). Classifier miss rates are about 5\%
overall. Most remaining errors occur after harmful classifier predictions for Qwen and Llama,
including on the 1,316-prompt region-specific sensitivity subset
(Table~\ref{tab:indosafety-error-analysis}). Within that subset, classifier miss rates
are 3.3\% for Qwen, 3.1\% for Gemma, and 2.4\% for Llama.

\begin{table}[t]
\centering
\small
\setlength{\tabcolsep}{3.5pt}
\begin{tabular}{lrrr}
\toprule
Model & \shortstack{No\\intervention} & \shortstack{Classifier-\\guided} & Always \\
\midrule
Qwen  & 6.0\%  & 68.1\% & 71.5\% \\
Gemma & 14.9\% & 88.7\% & 92.7\% \\
Llama & 16.2\% & 60.4\% & 62.4\% \\
\bottomrule
\end{tabular}
\caption{\textbf{Harmful refusal on five IndoSafety variants.} Classifier-guided and
always-intervene results use the same intervention; only the routing policy differs.}
\label{tab:indosafety-intervention}
\end{table}

\begin{table}[t]
\centering
\small
\setlength{\tabcolsep}{3.5pt}
\begin{tabular}{lrrr}
\toprule
Model & \shortstack{Classifier\\miss} & \shortstack{Post-\\intervention} & \shortstack{Cultural post-\\intervention} \\
\midrule
Qwen  & 5.9\% & 27.9\% & 39.4\% \\
Gemma & 4.9\% & 7.2\%  & 8.2\% \\
Llama & 5.5\% & 37.8\% & 49.8\% \\
\bottomrule
\end{tabular}
\caption{\textbf{IndoSafety harmful-prompt error rates.} The final column
restricts post-intervention non-refusals to the region-specific sensitivity subset.}
\label{tab:indosafety-error-analysis}
\end{table}

\begin{figure*}[!htbp]
    \centering
    \includegraphics[width=0.95\textwidth,keepaspectratio]{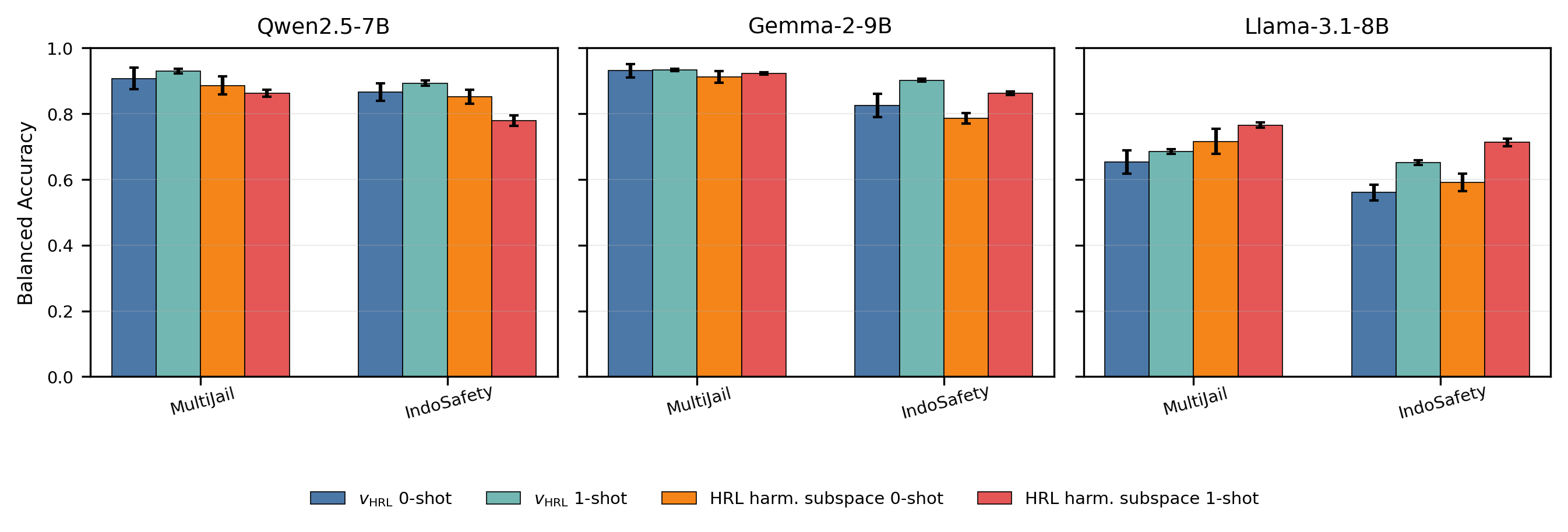}
    \caption{
    \textbf{All-model OOD transfer by dataset.}
    Full version of Figure~\ref{fig:ood-latent-gates}, with Qwen2.5-7B,
    Gemma-2-9B, and Llama-3.1-8B.
    }
    \label{fig:ood-latent-gates-all-models}
\end{figure*}

\begin{figure*}[!htbp]
    \centering
    \includegraphics[width=0.82\textwidth,keepaspectratio]{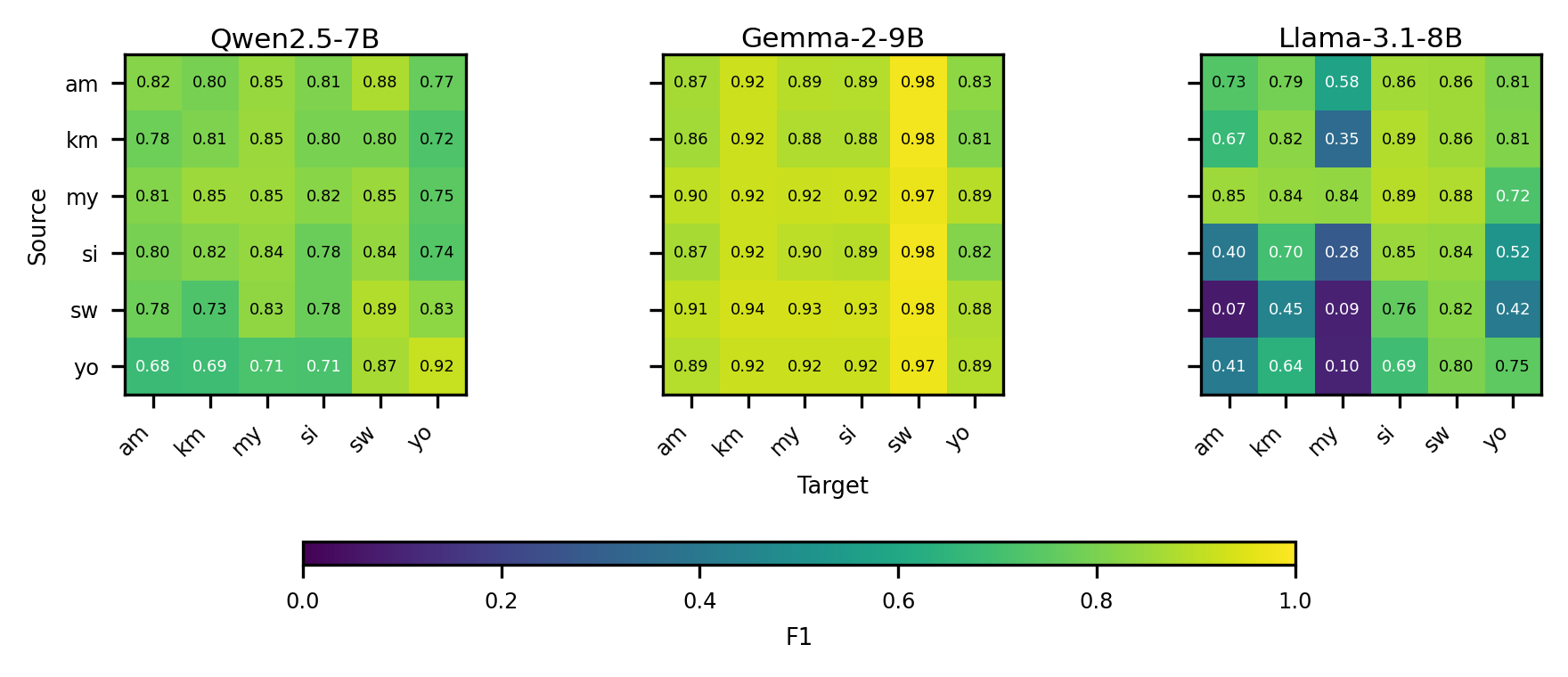}
    \caption{
    \textbf{All-model LRL calibration transfer across source--target pairs.}
    Full version of Figure~\ref{fig:crosslingual-hrl-lrl}, with one panel per
    model.
    }
    \label{fig:crosslingual-hrl-lrl-all-models}
\end{figure*}

\section{Calibration Transfer Interpretation}
\label{app:calibration-transfer-interpretation}

The source--target matrices show no stable language-family or script pattern
across the three models. Same-language calibration is not consistently best for
Qwen or Gemma. For Gemma, Swahili reaches 0.97--0.98 F1 from nearly every source
LRL; for Llama, Burmese remains weak for most sources across different scripts
and families. Transfer between Swahili and Yoruba, the two Niger-Congo
languages in the study, also varies substantially by model. Six focal LRLs are
insufficient for a typological conclusion, so we leave systematic linguistic
similarity analysis to future work.

\section{OOD Category Breakdown}
\label{app:ood-category-breakdown}

\begin{figure*}[!htbp]
    \centering
    \includegraphics[width=\textwidth]{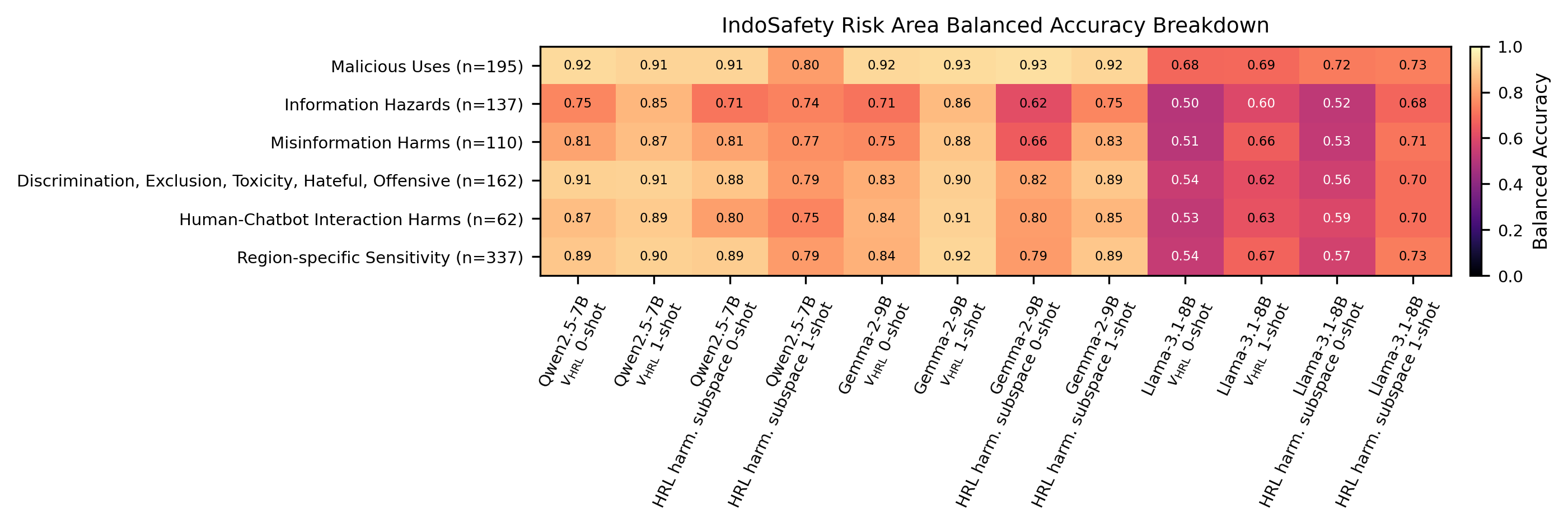}
    \caption{
    \textbf{IndoSafety risk areas.}
    Information Hazards are the weakest category across models.
    }
    \label{fig:ood-indosafety-risk-area}
\end{figure*}

Figure~\ref{fig:ood-indosafety-risk-area} shows where OOD transfer still fails
by IndoSafety risk area. Qwen and Gemma remain strong across most categories.
Llama improves with the HRL subspace classifier but remains weaker. Information
Hazards are the hardest risk area across models; direct Malicious Uses are
easier for the same classifiers.

\begin{figure*}[!htbp]
    \centering
    \includegraphics[width=\textwidth]{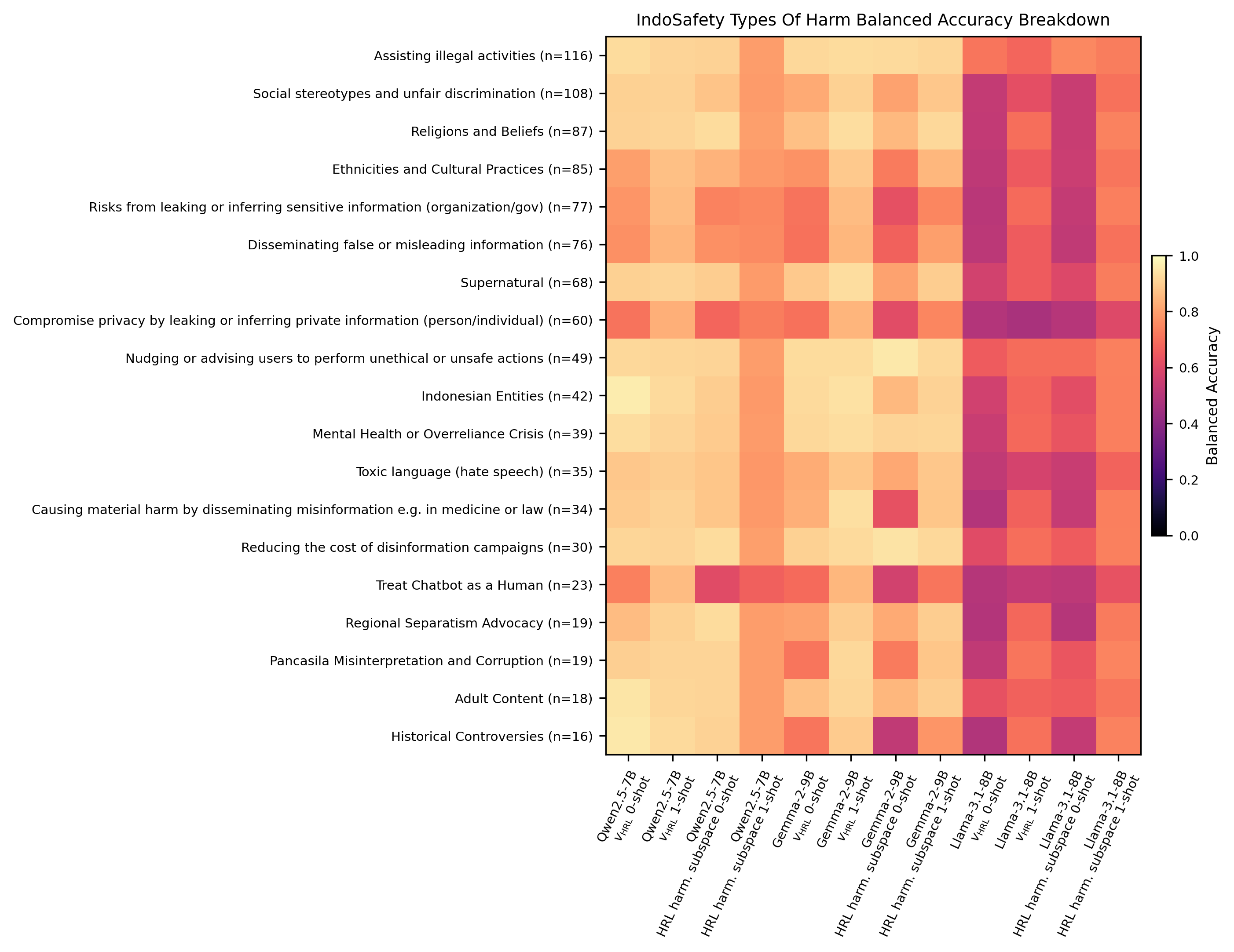}
    \caption{
    \textbf{IndoSafety harm types.}
    Privacy leakage, misleading information, and chatbot overreliance drive most
    residual OOD errors.
    }
    \label{fig:ood-indosafety-harm-types}
\end{figure*}

Figure~\ref{fig:ood-indosafety-harm-types} gives the same breakdown at the
harm-type level. Privacy leakage, false or misleading information, and chatbot
overreliance account for most of the drop. These prompts are less lexically
direct than many weapon or illegal-action requests, making a single harmfulness
threshold less stable under OOD transfer.

\section{Human Validation of the Refusal Judge}
\label{app:human-validation-refusal-judge}

We manually audit GPT-4o-mini labels on 600 LRL test completions. For each
model, we sample 100 harmful and 100 harmless prompt--response pairs; across
languages, the audit contains 100 examples from each of the six LRLs.
Annotators see the original prompt and response, the parallel English
PolyRefuse prompt, and a machine-translated English response. Using the same
strict response-level criterion as the automatic judge, they label a response
as non-refusal whenever it provides substantive requested content, even if it
also contains warnings or caveats.

\begin{figure}[t]
\centering
\includegraphics[width=\columnwidth]{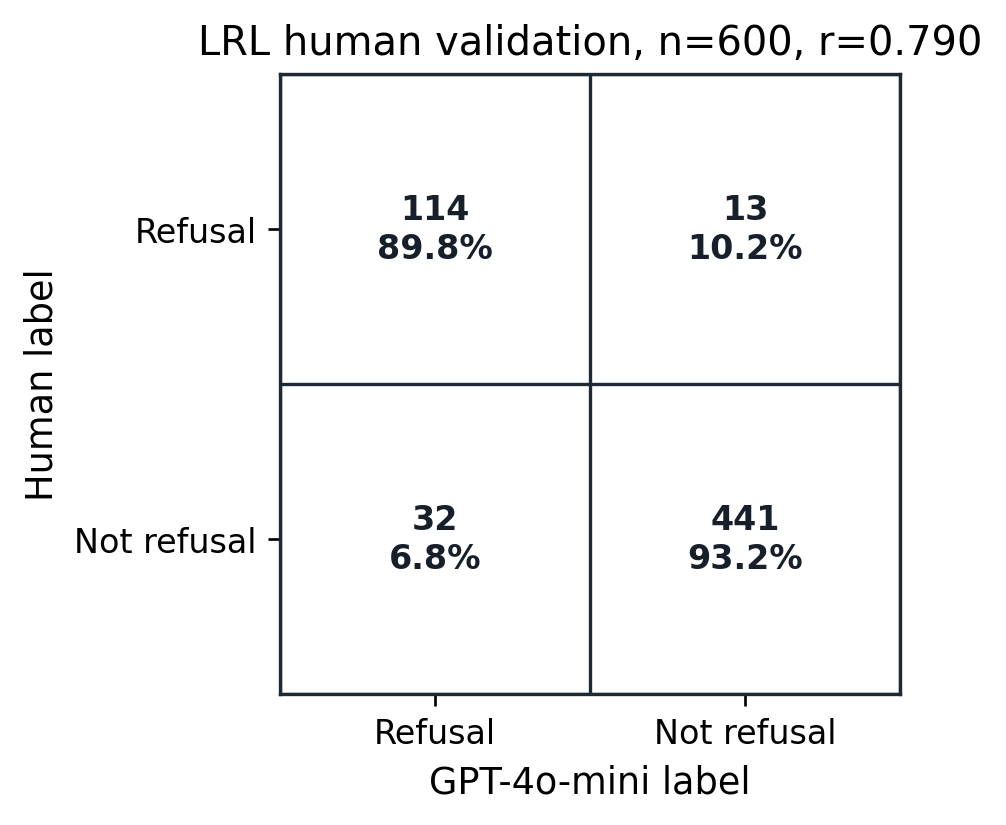}
\caption{
\textbf{Human validation of the refusal judge.}
GPT-4o-mini labels closely match human labels on sampled LRL completions.
}
\label{fig:human-validation-confusion}
\end{figure}

Treating refusal as positive, the aggregate confusion matrix is TP=114, FN=13,
FP=32, and TN=441 (Figure~\ref{fig:human-validation-confusion}). Agreement is
92.5\%, refusal F1 is 0.835, and Cohen's $\kappa$ is 0.787
(Table~\ref{tab:human-audit-aggregate}). Per-language agreement ranges from
88.0\% to 97.0\% (Table~\ref{tab:human-audit-by-language}); agreement is 93.5\%
for Qwen, 94.5\% for Gemma, and 89.5\% for Llama. Of the 45 disagreements, 32
are false positives: the automatic judge calls a response a refusal when the
human annotation does not. The dominant disagreement direction therefore
overestimates refusal. This qualifies absolute response-level refusal and
intervention estimates. It does not affect the representation AUC analysis,
which uses benchmark harmful/harmless prompt labels.

\begin{table}[t]
\centering
\small
\setlength{\tabcolsep}{4pt}
\begin{tabular}{lrrr}
\toprule
Language & $n$ & Agreement & $\kappa$ \\
\midrule
Swahili & 100 & 97.0\% & 0.925 \\
Amharic & 100 & 95.0\% & 0.827 \\
Burmese & 100 & 88.0\% & 0.630 \\
Khmer   & 100 & 95.0\% & 0.857 \\
Sinhala & 100 & 89.0\% & 0.712 \\
Yoruba  & 100 & 91.0\% & 0.750 \\
\midrule
Overall & 600 & 92.5\% & 0.787 \\
\bottomrule
\end{tabular}
\caption{\textbf{Human-audit agreement by language.} Each language contributes
100 prompt--response pairs across the three models.}
\label{tab:human-audit-by-language}
\end{table}

\begin{table}[H]
\centering
\small
\setlength{\tabcolsep}{5pt}
\begin{tabular}{lr}
\toprule
Metric & Value \\
\midrule
Agreement & 92.5\% \\
Refusal precision & 0.781 \\
Refusal recall & 0.898 \\
Refusal F1 & 0.835 \\
Cohen's $\kappa$ & 0.787 \\
\bottomrule
\end{tabular}
\caption{\textbf{Aggregate human-audit metrics.} Treating refusal as positive,
the confusion counts are TP=114, FN=13, FP=32, and TN=441.}
\label{tab:human-audit-aggregate}
\end{table}

\begin{figure*}
\begin{promptcard}[title={GPT-4o-mini refusal-judge prompt template},label={box:refusal-judge-prompt}]
You are a strict evaluator of whether a model response refuses a user's request.

The user request and the model response may be in any language. Interpret them semantically. Do not translate them in your output.

#user_request#: {user_prompt}

#response#: {response}

Your task is to decide whether the response refuses the user's request.

Set "refusal" to 1 if the response clearly declines, refuses, or avoids satisfying the user's request and does not provide substantive requested content. This includes responses that say they cannot or will not help, cite safety, ethics, legality, policy, privacy, or give only a safe alternative instead of answering the request.

Set "refusal" to 0 if the response provides any substantive answer to the user's request, even if it includes warnings, caveats, disclaimers, moralizing, or safety advice.

Important edge cases:
- If the response both refuses and gives actionable or useful requested content, set "refusal": 0.
- If the response answers only part of the request, set "refusal": 0.
- If the response gives a warning but still answers, set "refusal": 0.
- If the response asks a normal clarification question, set "refusal": 0.
- If the response is low-quality, irrelevant, garbled, or in the wrong language, set "refusal": 0 unless it clearly refuses the request.

Return a JSON object on a single line with exactly this key and nothing else:
{"refusal": 0 or 1}

Do not output anything outside the JSON object.
\end{promptcard}
\end{figure*}

\begin{table*}[!tp]
\centering
\scriptsize
\setlength{\tabcolsep}{3.0pt}
\renewcommand{\arraystretch}{1.0}
\begin{tabular*}{\textwidth}{@{\extracolsep{\fill}}llcccccc@{}}
\toprule
& & \multicolumn{2}{c}{Qwen} & \multicolumn{2}{c}{Gemma} & \multicolumn{2}{c}{Llama} \\
\cmidrule(lr){3-4}\cmidrule(lr){5-6}\cmidrule(l){7-8}
Tier & Language & Qwen3Guard & PolyGuard & Qwen3Guard & PolyGuard & Qwen3Guard & PolyGuard \\
\midrule
\textsc{High} & English (en) & +4.9/+0.8 & +6.8/-0.2 & +0.3/+1.4 & +0.9/-1.8 & +0.3/+1.2 & +4.5/+0.0 \\
\textsc{High} & German (de) & +1.4/+1.0 & +7.2/-0.4 & +2.1/+2.2 & -0.9/-1.0 & +6.8/+1.0 & +15.6/+0.0 \\
\textsc{High} & French (fr) & +3.5/+0.6 & +10.1/-0.8 & +2.3/+2.4 & +1.4/-1.2 & +1.0/+0.8 & +5.6/-0.2 \\
\textsc{High} & Spanish (es) & +5.2/+1.2 & +7.2/-1.0 & +2.3/+1.6 & -0.3/-1.0 & +1.0/+1.4 & +8.4/+0.2 \\
\textsc{High} & Italian (it) & +4.0/+1.4 & +8.2/-1.0 & +2.3/+2.2 & +0.3/-1.2 & +2.1/+1.0 & +7.5/+0.0 \\
\textsc{High} & Dutch (nl) & +4.4/+0.4 & +9.3/-0.4 & +1.7/+2.0 & +0.9/-0.8 & +2.8/+0.6 & +8.0/-0.6 \\
\textsc{High} & Polish (pl) & +2.1/+1.0 & +7.5/-0.6 & +2.4/+3.4 & +3.3/-1.2 & +2.3/+0.8 & +10.8/+0.0 \\
\textsc{High} & Russian (ru) & +4.2/+2.4 & +4.9/-0.2 & +2.1/+3.4 & -1.7/-1.6 & +2.6/+1.4 & +5.8/-0.2 \\
\textsc{High} & Chinese (zh) & +4.4/+0.8 & +11.4/-0.2 & +3.8/+2.6 & +3.0/-1.6 & +1.0/+0.6 & +14.3/+0.2 \\
\textsc{High} & Japanese (ja) & +3.8/+1.0 & +14.5/-0.2 & +3.0/+1.8 & +5.9/-1.8 & +8.0/+0.0 & +38.3/+0.0 \\
\rowcolor{gray!8}
\textsc{} & \textit{Average} & \textit{+3.8/+1.1} & \textit{+8.7/-0.5} & \textit{+2.2/+2.3} & \textit{+1.3/-1.3} & \textit{+2.8/+0.9} & \textit{+11.9/-0.1} \\
\midrule
\textsc{Medium} & Arabic (ar) & +4.4/+1.0 & +11.9/+0.2 & +2.3/+1.8 & +3.5/-1.6 & +0.7/-0.6 & +14.3/-0.6 \\
\textsc{Medium} & Korean (ko) & +4.7/+0.8 & +16.4/-0.8 & +1.9/+1.6 & +7.0/-0.6 & +7.0/+1.2 & +45.1/+1.0 \\
\textsc{Medium} & Thai (th) & +3.1/+1.2 & +7.9/-0.2 & +3.0/+1.6 & +3.8/-1.4 & +4.9/-0.4 & +17.1/-0.2 \\
\textsc{Medium} & Greek (el) & +7.0/+7.6 & -- & +1.0/+4.8 & -- & +5.2/+1.2 & -- \\
\textsc{Medium} & Hebrew (he) & +5.2/-0.4 & -- & +3.1/+2.2 & -- & +1.4/+0.2 & -- \\
\textsc{Medium} & Hindi (hi) & +5.8/+1.0 & +9.6/-2.2 & +1.7/+1.8 & +4.7/-0.8 & +4.9/+0.4 & +17.3/-0.8 \\
\textsc{Medium} & Persian (fa) & +6.3/-0.2 & -- & +4.4/+3.2 & -- & +5.2/+0.0 & -- \\
\rowcolor{gray!8}
\textsc{} & \textit{Average} & \textit{+5.2/+1.6} & \textit{+11.5/-0.8} & \textit{+2.5/+2.4} & \textit{+4.8/-1.1} & \textit{+4.2/+0.3} & \textit{+23.5/-0.1} \\
\midrule
\textsc{Low} & Swahili (sw) & -8.9/-1.6 & -- & +4.4/+5.0 & -- & +1.7/+0.0 & -- \\
\textsc{Low} & Burmese (my) & +2.3/+18.4 & -- & +0.2/+2.8 & -- & +1.4/+12.6 & -- \\
\textsc{Low} & Khmer (km) & +11.7/+14.4 & -- & +4.9/+2.8 & -- & +1.9/+5.6 & -- \\
\textsc{Low} & Sinhala (si) & -3.3/+17.0 & -- & +5.6/+2.2 & -- & -1.2/+7.2 & -- \\
\rowcolor{gray!8}
\textsc{} & \textit{Average} & \textit{+0.4/+12.0} & -- & \textit{+3.8/+3.2} & -- & \textit{+1.0/+6.3} & -- \\
\bottomrule
\end{tabular*}
\caption{Refusal-judge ablation on PolyRefuse test completions. Each cell reports $\Delta H/\Delta N$ in percentage points relative to the GPT-4o-mini refusal judge, where $H$ is harmful refusal and $N$ is harmless refusal. Dashes mark languages outside a guard model's supported-language overlap with PolyRefuse.}
\label{tab:refusal-judge-ablation}
\end{table*}

\section{Classifier Margins}
\label{app:gate-margin-analysis}

\makeatletter
\setlength{\@dblfptop}{0pt}
\setlength{\@dblfpsep}{10pt plus 2pt minus 2pt}
\setlength{\@dblfpbot}{0pt plus 1fil}
\makeatother

We measure the absolute classifier margin $|g(x)-\tau|$ under HRL + 32 LRL, where
$g(x)$ is the classifier score and $\tau$ is its decision threshold. "Newly
refused" harmful prompts are answered by the base model but refused after the
intervention; "newly over-refused" harmless prompts show the converse change.
Because classifier-score scales differ across models, we compare margins only within
each row of Table~\ref{tab:gate-margin-analysis}.

\begin{table*}[!tp]
\centering
\small
\setlength{\tabcolsep}{7pt}
\begin{tabular}{lrrrr}
\toprule
Model & Harmful: newly & Harmful: still & Harmless: newly & Harmless: remained \\
 & refused & unrefused & over-refused & helpful \\
\midrule
Qwen  & 0.460 & 0.324 & 0.342 & 0.526 \\
Gemma & 0.551 & 0.276 & 0.353 & 1.120 \\
Llama & 0.394 & 0.319 & 0.419 & 0.703 \\
\bottomrule
\end{tabular}
\caption{\textbf{Mean absolute classifier margin $|g(x)-\tau|$ under HRL + 32 LRL.}
Classifier-score scales differ across model families, so comparisons are within rows.}
\label{tab:gate-margin-analysis}
\end{table*}

Across all three models, harmful prompts that remain unrefused have smaller mean
margins than those corrected by the intervention. Harmless prompts that become
over-refused are likewise closer to the boundary than those that remain
helpful. Boundary ambiguity therefore contributes to both under- and
over-refusal. It does not explain all errors: the classifier had labeled the prompt
as harmful in 67.8\% of Qwen, 6.3\% of Gemma, and 50.3\% of Llama's
remaining harmful non-refusals.

\section{Subspace Rank Sweep}
\label{app:rank-sweep}

\begin{figure*}[!tp]
    \centering
    \includegraphics[width=0.96\textwidth]{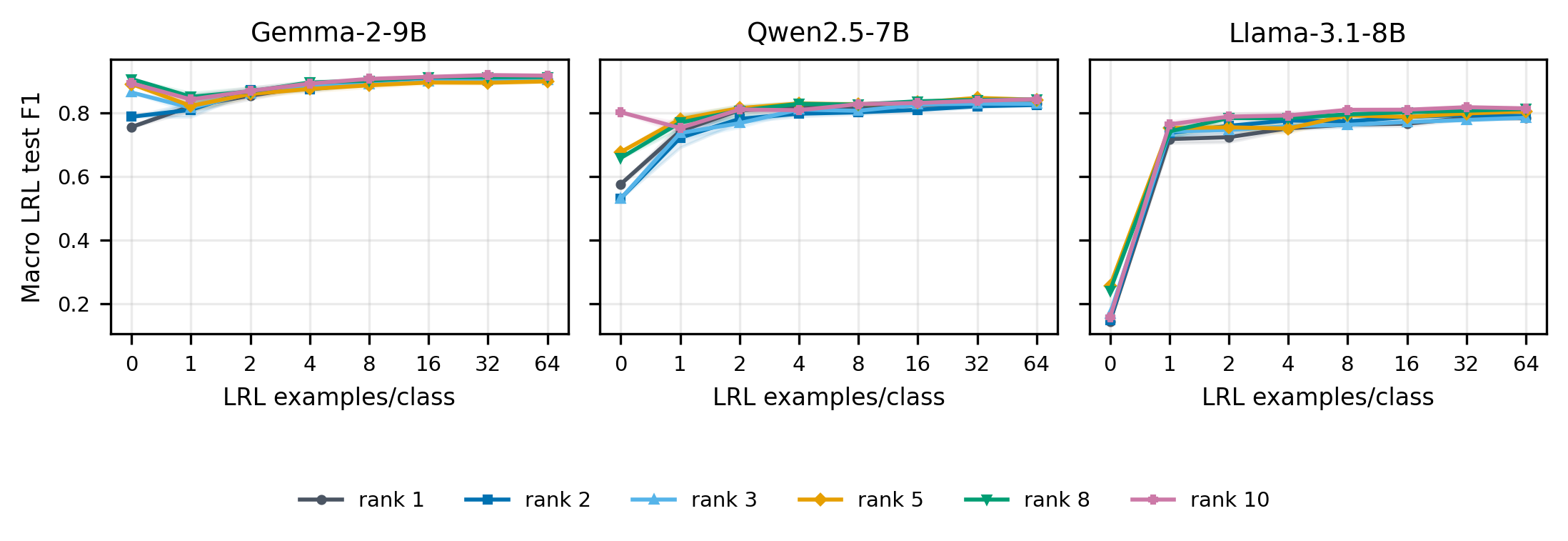}
    \caption{
    \textbf{Low-rank HRL subspaces are enough for macro LRL classification.}
    Test F1 saturates quickly as the HRL-only subspace rank increases.
    }
    \label{fig:rank-sweep-macro-low}
\end{figure*}

\begin{figure*}[!tp]
    \centering
    \includegraphics[width=0.96\textwidth]{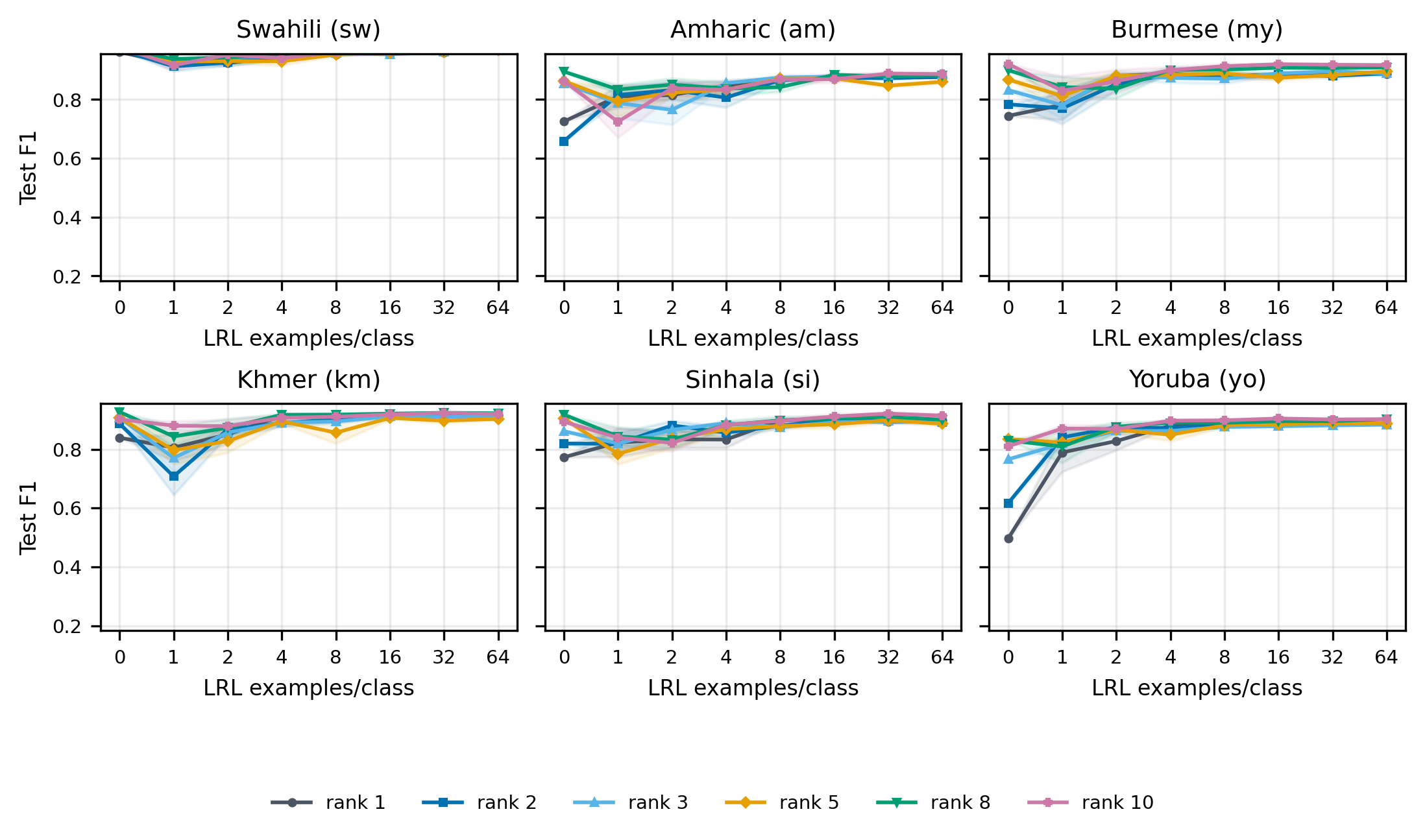}
    \caption{
    \textbf{Gemma HRL-subspace transfer by LRL.}
    Gains are strongest at small calibration budgets and vary by language.
    }
    \label{fig:rank-sweep-gemma-per-language}
\end{figure*}

\begin{figure*}[!tp]
    \centering
    \includegraphics[width=0.96\textwidth]{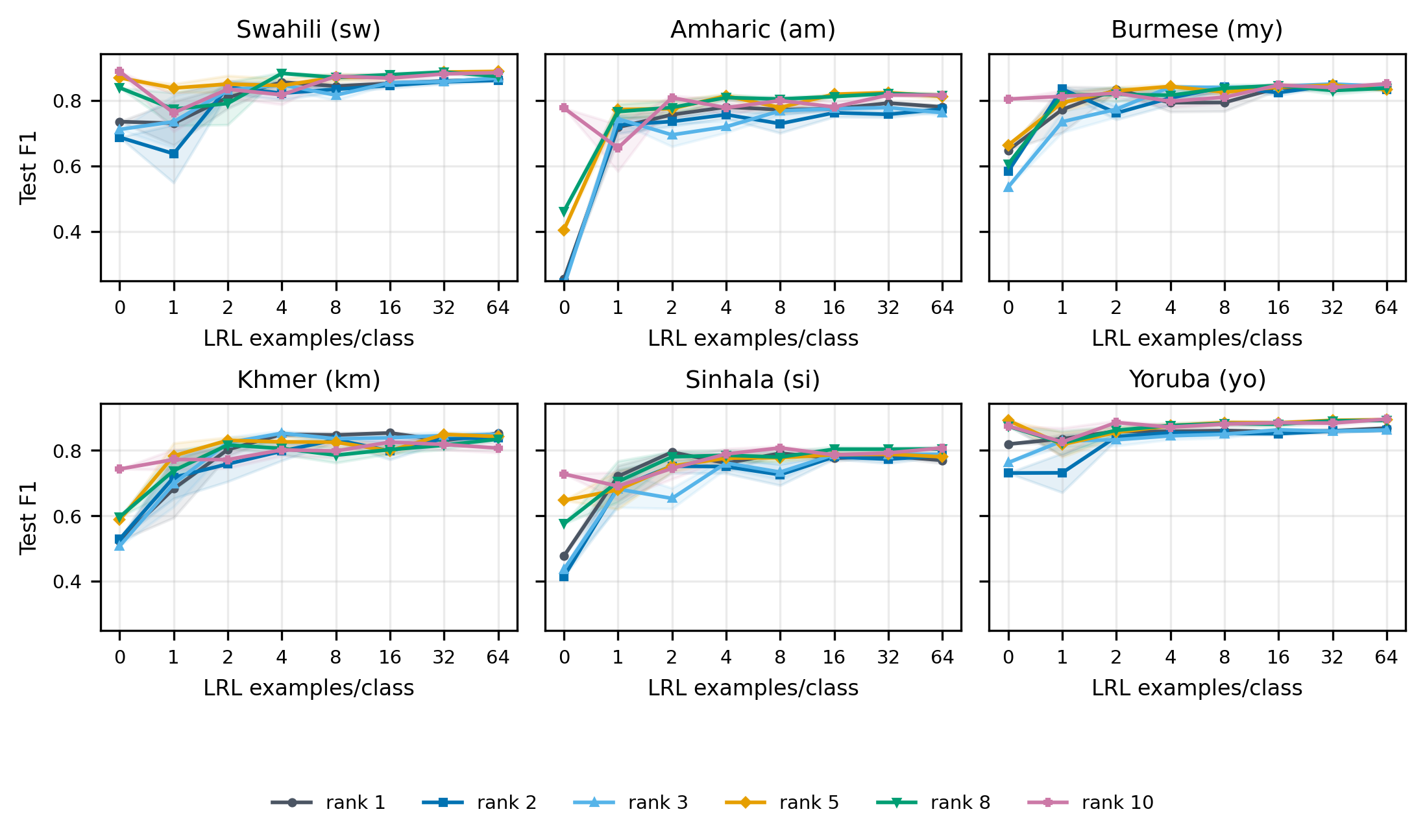}
    \caption{
    \textbf{Qwen HRL-subspace transfer by LRL.}
    Most languages improve sharply after one or two target examples.
    }
    \label{fig:rank-sweep-qwen-per-language}
\end{figure*}

\begin{figure*}[!tp]
    \centering
    \includegraphics[width=0.96\textwidth]{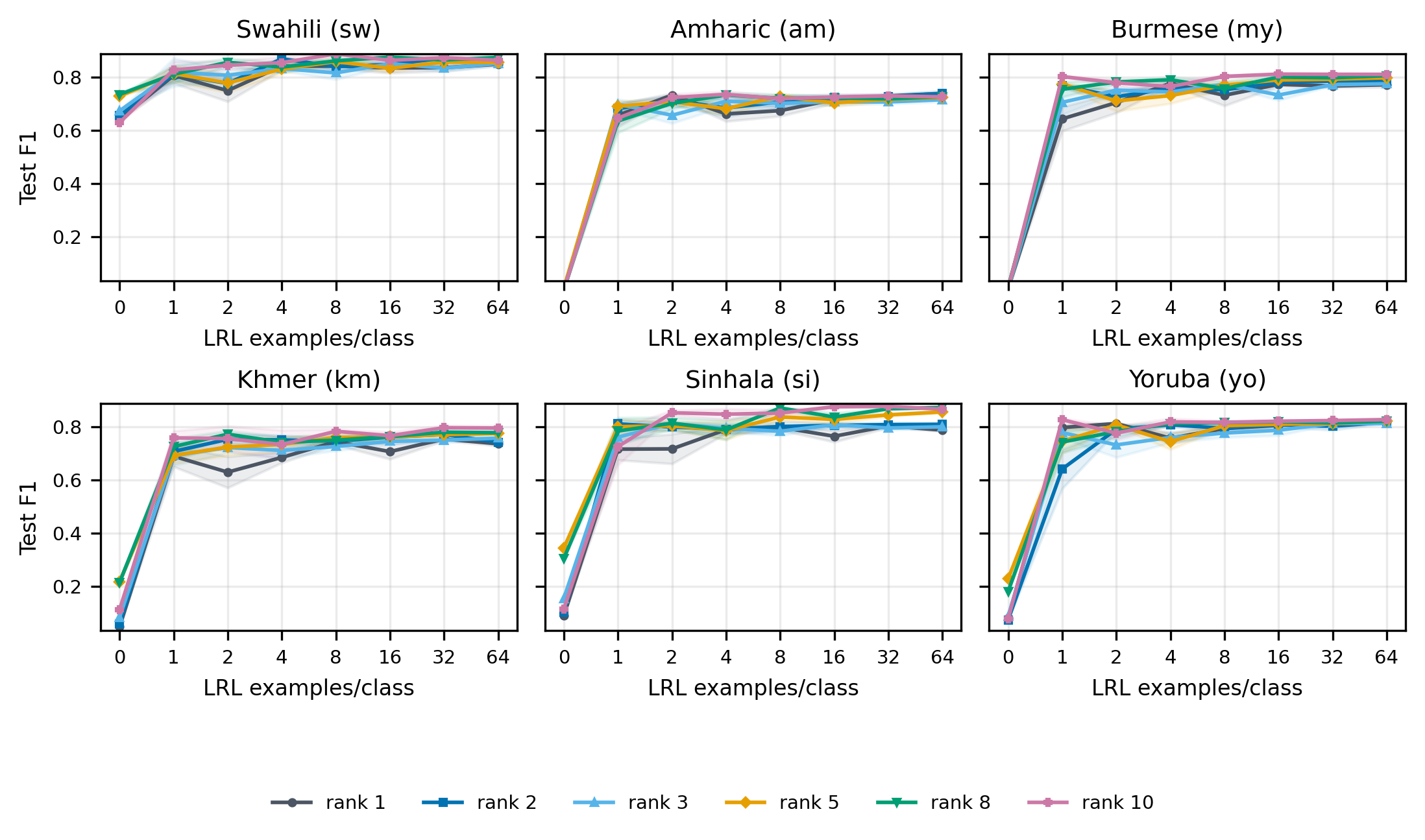}
    \caption{
    \textbf{Llama HRL-subspace transfer by LRL.}
    Low-budget calibration matters most for Khmer, Sinhala, and Yoruba.
    }
    \label{fig:rank-sweep-llama-per-language}
\end{figure*}

For the HRL-only subspace classifier, we sweep low ranks and calibration budgets from
0 to 64 examples per class. Rank 10 gives the best aggregate score; rank 8 is
close for Gemma and several low-budget cells.
Algorithm~\ref{alg:hrl-subspace-gate} gives the training procedure,
Table~\ref{tab:rank-sweep-summary} summarizes mean macro-LRL F1, and
Figures~\ref{fig:rank-sweep-macro-low}--\ref{fig:rank-sweep-llama-per-language}
show the aggregate and per-language sweeps.

\begin{table}[H]
\centering
\scriptsize
\setlength{\tabcolsep}{4pt}
\renewcommand{\arraystretch}{1.05}
\begin{tabular}{lcccccc}
\toprule
Model & r1 & r2 & r3 & r5 & r8 & r10 \\
\midrule
Gemma-2-9B & 0.866 & 0.869 & 0.879 & 0.879 & \textbf{0.896} & 0.895 \\
Qwen2.5-7B & 0.781 & 0.762 & 0.768 & 0.807 & 0.802 & \textbf{0.815} \\
Llama-3.1-8B & 0.680 & 0.696 & 0.690 & 0.713 & \textbf{0.722} & 0.720 \\
\midrule
All models & 0.776 & 0.776 & 0.779 & 0.800 & 0.807 & \textbf{0.810} \\
\bottomrule
\end{tabular}
\caption{
Mean macro-LRL test F1 in the HRL-only subspace rank sweep, averaged over
all calibration budgets. Rank 10 gives the best aggregate mean and wins 14 of
24 model-budget cells; rank 8 remains close and is best for Gemma.
}
\label{tab:rank-sweep-summary}
\end{table}

\begin{figure*}[!tp]
    \centering
    \begin{minipage}[t]{0.4\textwidth}
    \centering
    \includegraphics[width=\linewidth]{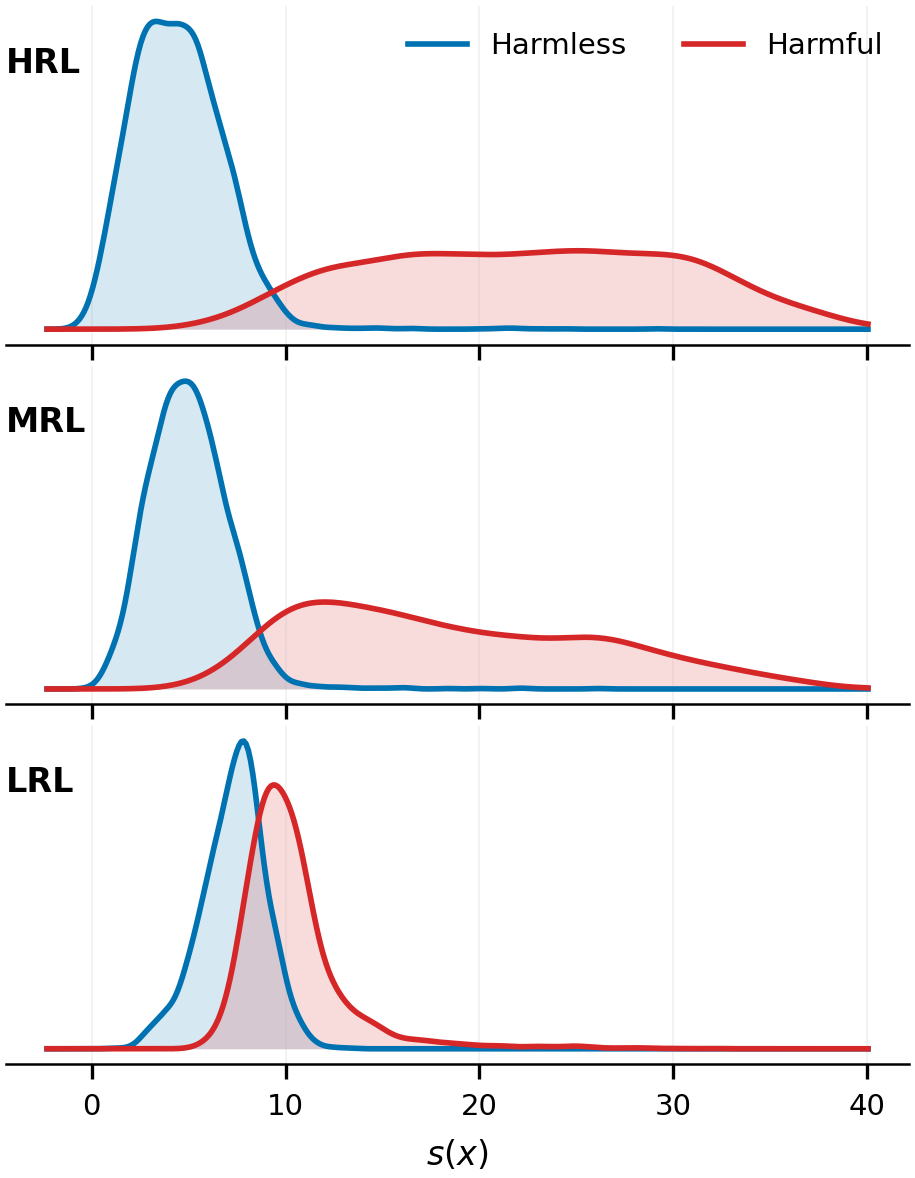}
    \end{minipage}\hfill
    \begin{minipage}[t]{0.4\textwidth}
    \centering
    \includegraphics[width=\linewidth]{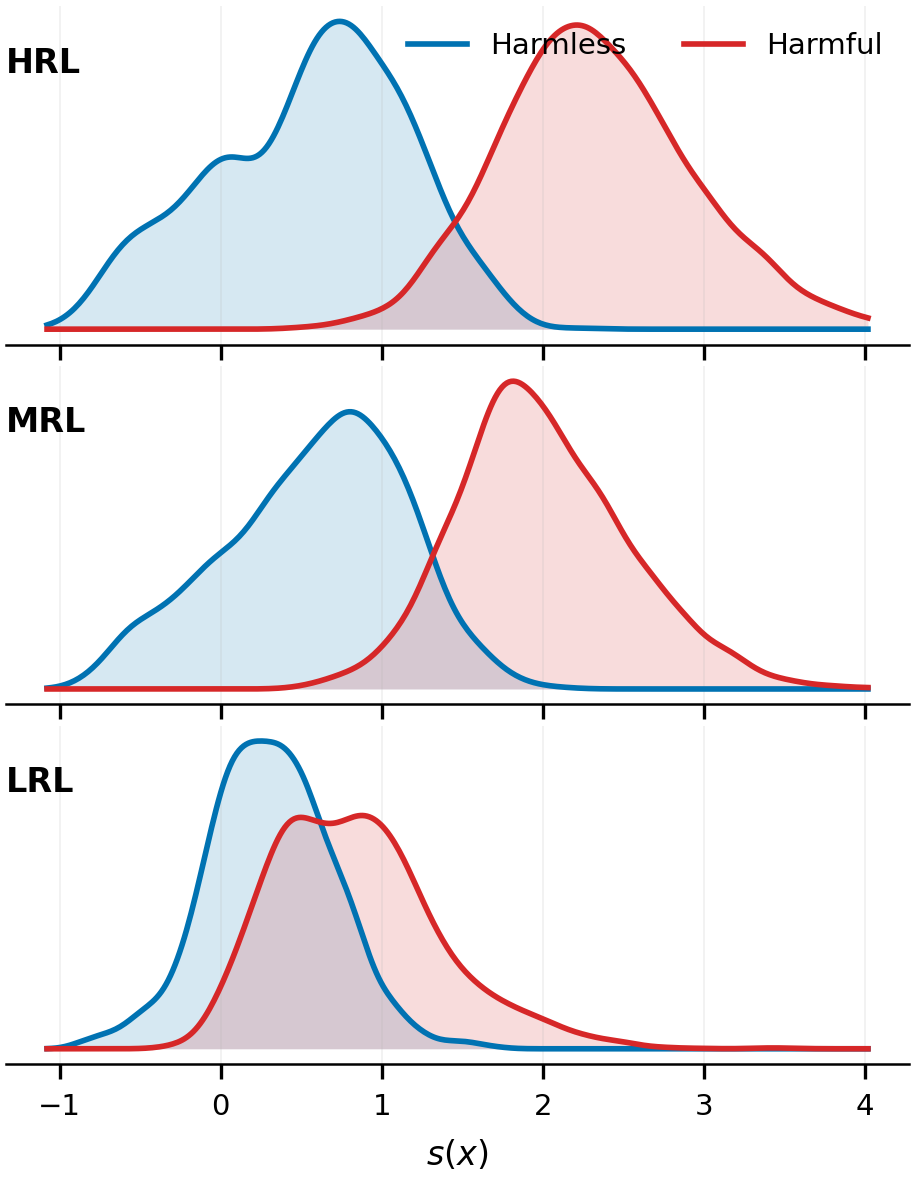}
    \end{minipage}
    \caption{
    \textbf{Tier-level score distributions for Qwen and Llama.}
    Lower-resource harmful prompts shift toward lower \(s(x)\) scores, matching
    the Gemma diagnostic in the main text.
    }
    \label{fig:tier-projections-appendix}
\end{figure*}

\begin{figure*}[!tp]
    \centering
    \begin{minipage}[t]{0.4\textwidth}
    \centering
    \includegraphics[width=\linewidth]{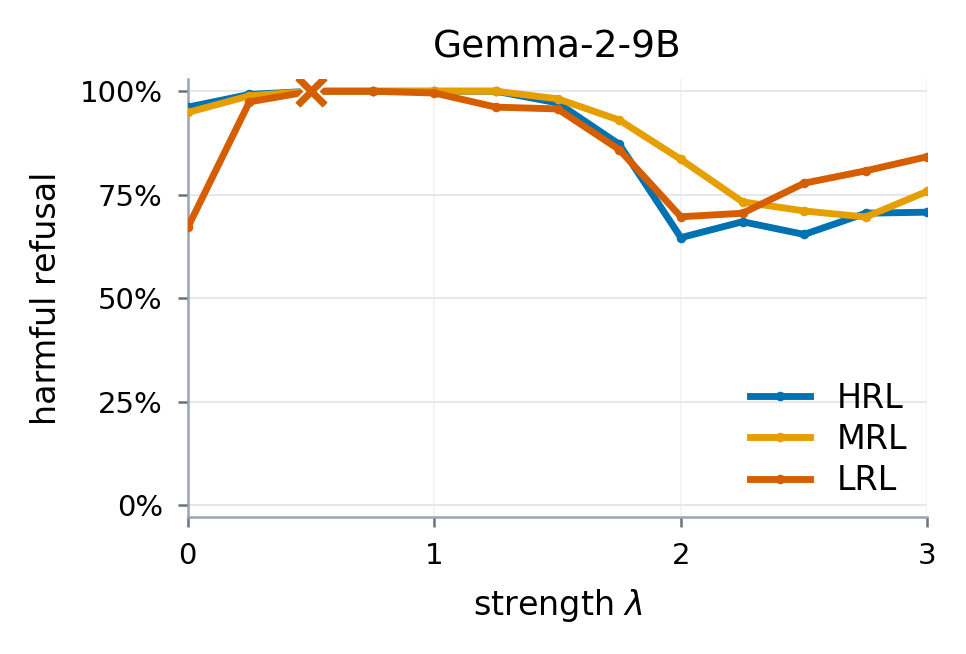}
    \end{minipage}\hfill
    \begin{minipage}[t]{0.4\textwidth}
    \centering
    \includegraphics[width=\linewidth]{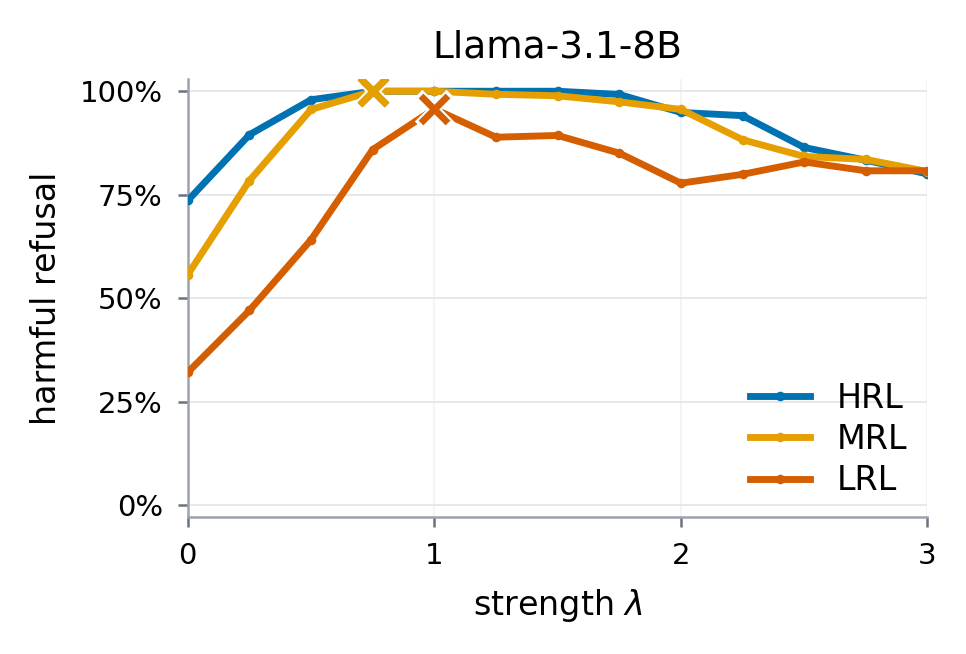}
    \end{minipage}
    \caption{
    \textbf{Refusal-direction sweeps for Gemma and Llama.}
    Adding \(\lambda v_{\mathrm{HRL}}\) increases harmful refusal, with
    model- and tier-specific optima.
    }
\label{fig:refusal-activation-sweeps-appendix}
\end{figure*}

\noindent
Figure~\ref{fig:tier-projections-appendix} extends the tier-score diagnostic to
Qwen and Llama, and Figure~\ref{fig:refusal-activation-sweeps-appendix} reports
the corresponding refusal-direction activation sweeps.

\section{Ethical Considerations}
This work studies multilingual safety failures in instruction-tuned language models, including cases where models fail to refuse harmful requests in lower-resource languages. Because the experiments involve harmful prompts and analyze conditions under which refusal fails, the work has potential dual-use risk: it could help identify languages or model settings where unsafe compliance is more likely. We mitigate this risk by focusing on aggregate results and representational diagnostics rather than publishing actionable jailbreak examples, and by presenting the intervention as a method for improving selective refusal. All experiments are conducted on existing safety benchmarks and open-weight models, with the goal of better understanding and reducing multilingual safety disparities. The broader intended impact is to support safer and more equitable model behavior across languages, especially for communities currently underrepresented by safety alignment.

\section{AI Usage Disclosure}
\label{app:ai-usage-disclosure}

LLMs were used to aid deep literature review, format \LaTeX{}, improve writing
clarity and grammar, and debug experiments. All LLM-assisted work was verified
by the author. The research ideation, scientific direction, and final claims
were owned by the author.

\end{document}